\documentclass[lettersize,journal]{IEEEtran}
\usepackage{amsmath,amsfonts}
\usepackage{algorithmic}
\usepackage{algorithm}
\usepackage{array}
\usepackage[caption=false,font=normalsize,labelfont=sf,textfont=sf]{subfig}
\usepackage{textcomp}
\usepackage{stfloats}
\usepackage{url}
\usepackage{verbatim}
\usepackage{graphicx}
\usepackage{cite}
\usepackage[table]{xcolor}
\usepackage[pagebackref,breaklinks,colorlinks]{hyperref}
\usepackage{tabularx}
\usepackage{booktabs}
\usepackage{cleveref}
\usepackage{tikz}

\usepackage{multirow}
\usepackage{pifont}
\usepackage{caption}
\usepackage{subcaption}
\newcommand{\cmark}{\ding{51}} 
\newcommand{\xmark}{\ding{55}} 
\newcommand{\best}[1]{\cellcolor{red!20}#1}
\newcommand{\second}[1]{\cellcolor{blue!18}#1}

\begin{document}

\title{Hybrid-PCGC: Hybrid Point Cloud Geometry Compression Bridging Generalization-based Model and Implicit Neural Representation}

\author{Wenjie Huang$^{*}$, Qi Yang$^{*}$, Shuting Xia, He Huang, Cixiao Zhang, Yiling Xu$^{\dagger}$, and Zhu Li
\thanks{$^{*}$Equal contribution. $^{\dagger}$Corresponding author: Yiling Xu.}
\thanks{An earlier preprint version of this work is available on arXiv~\cite{hybridpcgc_arxiv}. The present manuscript has been substantially revised and extended.}
\thanks{Wenjie Huang, Shuting Xia, He Huang, Cixiao Zhang, and Yiling Xu are with the Cooperative Media Network Innovation Center, Shanghai Jiao Tong University, Shanghai 200240, China. E-mail: 
huangwenjie2023@sjtu.edu.cn;
xiashuting@sjtu.edu.cn;
huanghe0429@sjtu.edu.cn; 
cixiaozhang@sjtu.edu.cn;
yl.xu@sjtu.edu.cn.}%
\thanks{Qi Yang and Zhu Li are with the Department of Computer Science and Electrical Engineering, University of Missouri--Kansas City, Kansas City, MI 64110 USA. E-mail:
qiyang@umkc.edu
lizhu@umkc.edu
}%
}

\markboth{Journal of \LaTeX\ Class Files,~Vol.~14, No.~8, August~2021}%
{Shell \MakeLowercase{\textit{et al.}}: A Sample Article Using IEEEtran.cls for IEEE Journals}

\maketitle
\begin{abstract}
Efficient point cloud geometry compression (PCGC) is fundamental to 3D sensing applications like autonomous driving and digital twins. Recently, hybrid compression methods have emerged to bridge the gap between generalization-based and Implicit Neural Representation (INR) paradigms, delivering stable performance across diverse data sources. However, existing hybrid schemes often rely on complex base architectures with rigid parameter-updating strategies, and their weight compression is typically decoupled from the training loop, leading to suboptimal adaptation efficiency and uncontrolled bitstream overhead. To overcome these limitations, this paper proposes Hybrid-PCGC, a novel and efficient hybrid PCGC framework. Rather than treating instance adaptation as post-hoc fine-tuning, Hybrid-PCGC formulates hybrid compression as a joint geometry-and-model bitrate allocation problem, where instance-specific weight offsets are optimized as transmitted, rate-constrained variables. Architecturally, we decouple dataset-agnostic prior modeling from data-specific updates via four collaborative components: a Frozen General Module (FGM), a Dual Trainable Adaptive Module (DTAM), an Adaptive Scope Controller (ASC), and a Supervised Offset Compression Module (SOCM). The FGM establishes a streamlined, high-capacity foundation for global geometric regularities. Guided by FGM features, the parameter-constrained DTAM performs instance adaptation, thereby providing a lightweight parameter space for data-specific updates. ASC selects the update range for each instance according to coding difficulty, while SOCM optimizes and compresses the corresponding weight offsets under differentiable entropy supervision within the training loop. Experimental results demonstrate that Hybrid-PCGC achieves an optimal balance between compression fidelity and bitrate, obtaining a 13.96\% / 16.33\% bitrate reduction over G-PCC on the lossless/lossy benchmark, with up to 12.33\% bitrate saving over SOTA hybrid method on challenging dense point clouds. Our project can be seen on \href{https://github.com/huangwenjie2023/hybrid-pcgc.git}{https://github.com/huangwenjie2023/hybrid-pcgc.git}.
\end{abstract}

\begin{IEEEkeywords}
3D, INR-based, Overfit, Instance-adapt, Hybrid, Point Cloud, Geometric Compression
\end{IEEEkeywords}

\begin{figure}[htp]
    \centering
    \includegraphics[width=0.95\linewidth]{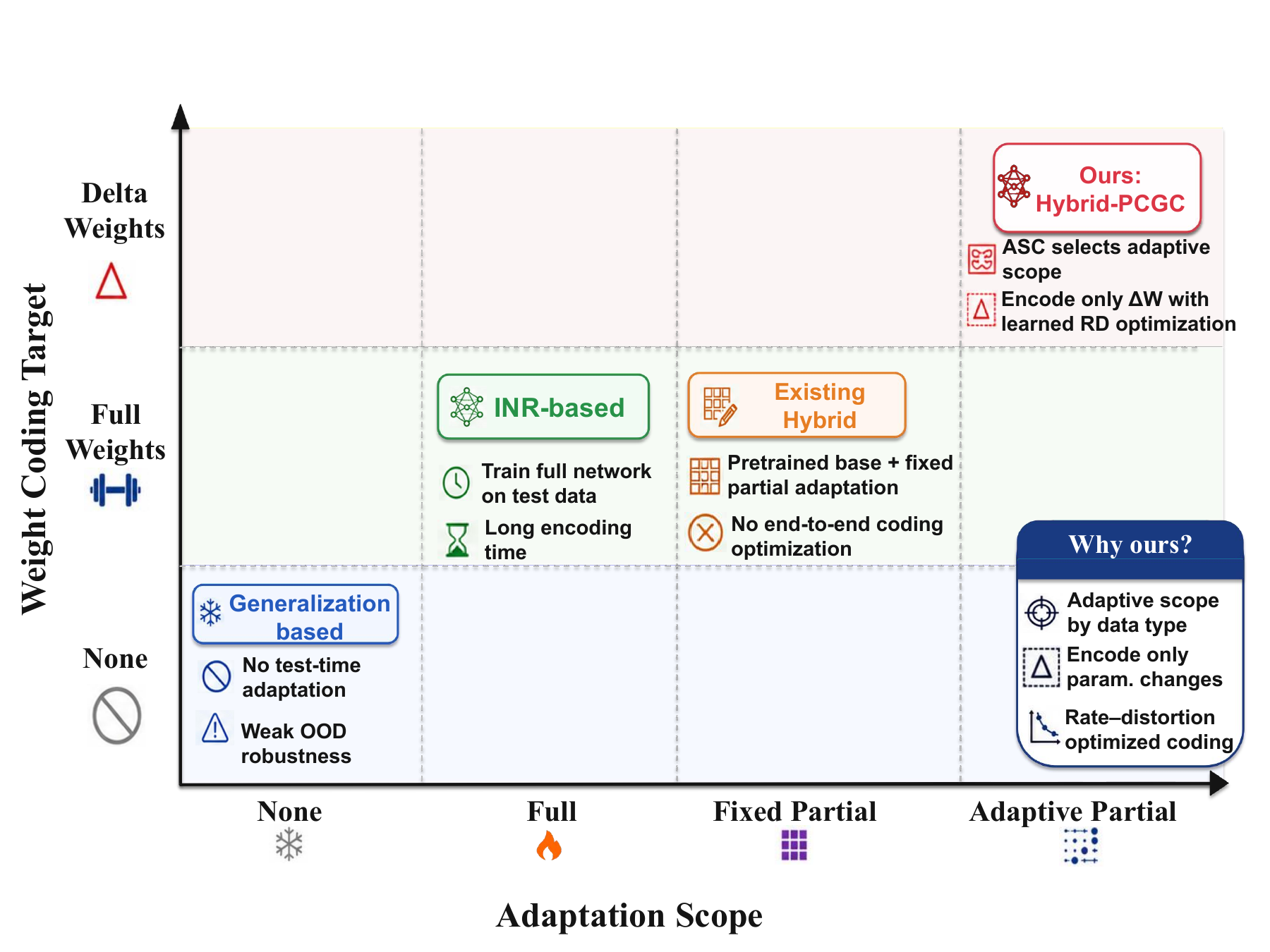}
    \caption{Principle comparison of generalization-based, INR-based, and our Hybrid approaches, highlighting the model components utilized during the pretrain, instance adaptation, and weight coding phases.}
    \label{fig:motivation}
    \vspace{-1.5em}
\end{figure}

\section{Introduction}
\label{sec:intro}

Driven by advancements in 3D sensing and computing power, point clouds have emerged as a foundational data format for numerous critical applications. They serve as the essential baseline for revolutionary fields such as autonomous driving~\cite{c13_topnet, c28_you2025}, Virtual Reality (VR), Augmented Reality (AR), and digital twins~\cite{c1_3dpcc_survey, c11_linr_pcgc}. A point cloud is commonly composed of both geometry and attributes: the geometric information defines the spatial structure and forms the crucial bedrock of the scene or object ~\cite{graphsim}, while attributes such as color or reflectance are strictly attached to the underlying geometry. 
Therefore, maintaining high-fidelity geometric accuracy is critical for high-quality rendering and downstream perception tasks \cite{MPED}. However, representing such precise geometric structures intrinsically generates an immense volume of data, posing severe storage and transmission challenges ~\cite{c2_dl_pcc_survey}. Therefore, efficient and accurate Point Cloud Geometry Compression (PCGC) deserves considerable attention, especially for high-stakes deployment scenarios that require stable coding performance across diverse data sources and varying point densities.

Traditional point cloud compression standards, represented by G-PCC~\cite{GPCC_method} and V-PCC~\cite{vpcc_paper}, establish comprehensive hand-crafted coding frameworks for different point cloud characteristics~\cite{c3_sparse_dense_pcc,c4_emerging_mpeg_pcc}. In general, G-PCC directly encodes 3D geometry by organizing point clouds with spatial structures such as octrees, while V-PCC converts point clouds into 2D patch sequences and leverages mature video coding tools for compression~\cite{V-PCC}. Related geometry-oriented studies further explore volumetric occupancy coding, run-length representation, and predictive RAHT for point cloud geometry compression~\cite{tip_volumetric_geometry_krivokuca2020,tip_runlength_tzamarias2022,tip_raht_vishwanath2025}. However, their reliance on manually specified priors makes it difficult to capture complex geometric distributions and content-adaptive rate-distortion behaviors~\cite{c5_vpcc_gpcc_overview}.

Subsequently, inspired by the success of learned image compression, AI-based point cloud compression methods have achieved significant bitstream savings compared to traditional methods~\cite{c2_dl_pcc_survey}. Different from traditional rule-based codecs, AI-based methods aim to learn compact latent representations and probability models from data, so that the encoder can better match the statistical regularities of point cloud geometry. Existing AI-based methods can be broadly divided into three categories: generalization-based, INR-based, and hybrid methods. Generalization-based methods, such as SparsePCGC~\cite{c6_sparse_tensor_pcc}, UniPCGC~\cite{c12_unipcgc}, OctAttention~\cite{c9_octattention}, and TopNet~\cite{c13_topnet}, 
train a generalized neural codec offline and then use the fixed network parameters to encode and decode target point clouds~\cite{c11_linr_pcgc,c20_choy2019}. By learning compact representations and probabilistic context models from large-scale training data, these methods exhibit strong predictive performance when applied to data with a distribution similar to the training data. However, their coding efficiency can degrade significantly when applied to Out-Of-Distribution (OOD) data, since the learned priors may no longer match the geometric statistics of unseen point clouds~\cite{c14_anypcc}.

INR-based methods provide another promising paradigm for point cloud geometry compression. Through instance-level network adaptation, these methods report more stable and distribution-agnostic compression performance across diverse data sources~\cite{c11_linr_pcgc,c20_choy2019,c22_hu2022,c23_isik2022,c24_pistilli2022,c25_ruan2024,tip_incp_ruan2026,c29_chen2021,c30_dupont2021}. Instead of depending primarily on a fixed dataset-level prior, they optimize an implicit representation for each target instance, allowing the network to learn the geometry distribution of the specific instance adequately. However, such methods have to capture prior context from scratch during the instance-adaptation stage, which makes the encoding process time-consuming. Moreover, the entire set of overfitted network parameters must be encoded into the bitstream, creating significant model overhead~\cite{c11_linr_pcgc}.


To bridge this gap, hybrid methods attempt to marry the generalization capabilities of learning-based backbones with the instance-adaptation of INR paradigms. An example is AnyPCC~\cite{c14_anypcc}, which introduces an Instance-Adaptive Fine-Tuning (IAFT) strategy. Built upon a generalized backbone, AnyPCC optimizes only a minimal subset of parameters within the final MLP layers to bypass the prohibitive encoding delays and model overhead of pure INR methods, followed by a decoupled CABAC~\cite{c57_CABAC} step for weight compression. Despite this progress, existing hybrid codecs still treat instance adaptation mainly as a fixed fine-tuning operation. This leaves two rate-distortion questions insufficiently addressed: how much adaptive model capacity should be allocated to each point cloud, and how many bits should be spent to transmit the corresponding parameter changes. A fixed update range cannot dynamically control which network weights participate in instance adaptation, while post-processing weight coding decouples model-offset compression from the differentiable training objective.

In this paper, we focus on the hybrid method paradigm and propose Hybrid-PCGC to address the aforementioned limitations, as depicted in Fig. \ref{fig:motivation}. The core idea is to regard the adaptive part of the model as a compressible representation whose capacity and bitrate must be jointly controlled with the geometry stream. Specifically, we architecturally propose four collaborative components: a Frozen General Module (\textbf{FGM}), a Dual Trainable Adaptive Module (\textbf{DTAM}), an Adaptive Scope Controller (\textbf{ASC}), and a Supervised Offset Compression Module (\textbf{SOCM}). FGM is equipped with a large parameter capacity to capture rich, dataset-agnostic prior knowledge, serving as the robust foundation of our framework.  To break the limitation of the network parameter updating scope of FGM, DTAM is introduced as a structurally elastic space used for instance-adaptation, guided directly by features from FGM. While DTAM closely resembles the network topology of FGM, it strictly constrains its parameter capacity. This architectural design allows the network to scale its active parameters based on scene complexity, ensuring that the core priors in FGM remain uncorrupted. ASC then controls the scope of instance adaptation according to the specific coding scenario, selecting the DTAM parameter subset that should be updated. Finally, to eliminate decoupled post-processing, SOCM supervises the selected weight offset $\Delta W$ as a latent geometric feature within a differentiable entropy estimation loop. This design enables the framework to adaptively estimate parameter distributions and derive optimal quantization step sizes online, strictly controlling the weight bitstream size according to scene demands. The main contributions of this paper are summarized as follows:
\begin{itemize}
    \item We formulate hybrid PCGC as a joint geometry-and-model bitrate optimization problem, where instance-specific weight offsets are treated as transmitted, rate-constrained variables rather than post-hoc fine-tuned parameters.
    \item We design FGM and DTAM to explicitly separate dataset-agnostic prior modeling from data-specific updates, providing a frozen general prior together with a compact adaptive parameter space for controllable instance refinement.
    \item We introduce ASC and SOCM to allocate the adaptive parameter scope and compress the corresponding delta weights under entropy supervision, closing the rate-distortion loop between geometry bits and model-offset bits.
    \item Experiments show that Hybrid-PCGC achieves consistent bitrate savings across heterogeneous point cloud benchmarks, reducing bitrate by 13.96\% over G-PCC on average and up to 12.33\% over AnyPCC on challenging dense point clouds.
\end{itemize}

\section{Related Work}
\label{sec:related_work}
\subsection{Traditional Compression Methods}
Traditional Point Cloud Compression (PCC) methods standardized by the Moving Picture Experts Group (MPEG) serve as mature rule-based anchors for both sparse and dense data~\cite{c3_sparse_dense_pcc,c4_emerging_mpeg_pcc,c5_vpcc_gpcc_overview}. Geometry-based Point Cloud Compression (G-PCC)~\cite{GPCC_method,c4_emerging_mpeg_pcc} organizes voxelized geometry with hierarchical spatial partitions, such as octrees, and predicts occupancy patterns from local contexts using arithmetic coding~\cite{c5_vpcc_gpcc_overview}. Other classical and transform-oriented geometry codecs explore volumetric representations, fast run-length coding, predictive RAHT, and graph-based dynamic coding~\cite{tip_volumetric_geometry_krivokuca2020,tip_runlength_tzamarias2022,tip_raht_vishwanath2025,tip_graph_dynamic_thanou2016}. Video-based Point Cloud Compression (V-PCC)~\cite{vpcc_paper,vpcc_software} instead projects 3D surfaces into packed 2D patches to leverage video codecs, with subsequent works improving motion prediction, rate control, and low-complexity coding-unit decisions~\cite{vpcc_paper,c5_vpcc_gpcc_overview,tip_motion_prediction_li2020,tip_rate_control_li2020,tip_low_complexity_gao2024}. These hand-crafted codecs remain robust and interpretable, but their fixed context rules motivate learned models for more adaptive geometry probability estimation.

\subsection{Generalization-based Methods}
Generalization-based methods utilize deep neural networks to model complex spatial relationships and learn reusable probability models, typically categorised by how they represent and aggregate geometric context: point-wise operations preserve irregular details directly, voxel or sparse-tensor representations convert geometry into occupancy grids for efficient convolutions, and octree-like structures organize occupied space hierarchically for progressive probability estimation~\cite{c2_dl_pcc_survey,c31_xu2025PCCobjective}. Early learned approaches explored convolution transforms and end-to-end optimization for lossy geometry compression~\cite{c16_quach2019,c52_lossy_e2e}. PCGCv2~\cite{c17_wang2021} and SparsePCGC~\cite{c6_sparse_tensor_pcc} introduced multiscale sparse-tensor frameworks to exploit local occupancy patterns and cross-scale dependencies, with SparsePCGC further processing only the Most-Probable Positively-Occupied Voxels (MP-POV) using a SparseCNN-based Occupancy Probability Approximation (SOPA) model. UniPCGC~\cite{c12_unipcgc} and Unicorn~\cite{c7_umc_geometry} extend this direction toward more versatile multiscale coding, aiming to support diverse point clouds and compression settings within unified frameworks. In parallel, octree-oriented methods including OctSqueeze~\cite{c10_octsqueeze}, VoxelContext-Net~\cite{c19_que2021}, PCGFormer~\cite{c49_pcgformer}, OctAttention~\cite{c9_octattention}, TopNet~\cite{c13_topnet}, EHEM~\cite{c21_song2023}, OctFormer~\cite{c26_cui2023}, and ECM-OPCC~\cite{c27_jin2024} improve entropy modeling by designing stronger contexts, attention mechanisms, or hierarchical prediction strategies over occupied nodes. Recent works further refine feature extraction and context modeling through feature residual learning, space-to-channel transformations, content-specific priors, low-latency coding designs, and LiDAR-oriented predictive models~\cite{c33_voxel_stc,tip_stc_liu2026,c45_geometric_prior,tip_hier_prior_li2024,c34_pointsoup,c41_hier_dist,c50_dense_maps,c51_muscle,c28_you2025,c38_lpcm,tip_interlpcm_sun2026,tip_sdt_li2026,c39_gaem,c40_redensification}. Rate-reconfigurable, perceptual, and task-aware learned PCC have also been explored to better adapt bitrate allocation or representation targets to diverse application requirements~\cite{tip_rate_reconfig_zhang2026,tip_sagpcc_zhou2025}, while inter-frame geometry coding and neural point cloud video streaming address dynamic and streaming settings~\cite{tip_interframe_akhtar2024,tip_ncs_zhang2025}.

\subsection{INR-based Methods}
Implicit Neural Representation (INR) based compression stems from learned image compression and neural fields. Hyperprior image codecs~\cite{VAE}, coordinate-based shape representations~\cite{c53_deepsdf}, and parameter-as-signal codecs such as COIN~\cite{c30_dupont2021} and NeRV~\cite{c29_chen2021} inspired point cloud codecs that optimize a coordinate network for each instance, with the resulting weights forming the compressed bitstream.

Hu et al.~\cite{c22_hu2022} introduced neural volumetric fields for geometry reconstruction, while LVAC~\cite{c23_isik2022} and related methods~\cite{c24_pistilli2022} used coordinate-based networks to compress attributes or generic signals. Ruan et al.~\cite{c25_ruan2024,tip_incp_ruan2026} learned coordinate fields for joint geometry and attribute reconstruction. For LiDAR, NeRI~\cite{c36_neri} represents range-image sequences with INRs to exploit temporal correlation, and recent work explores implicit compression via learnable activations in latent space~\cite{c37_einc_latent_activation}. To mitigate model overhead, LINR-PCGC~\cite{c11_linr_pcgc} introduced a Group of Point Cloud (GoPC) framework, overfitting and sharing a lightweight network across temporal frames with sequential initialization. These methods improve distribution robustness, but online per-instance training and encoded network weights remain expensive for lossless or low-latency coding~\cite{c14_anypcc,c11_linr_pcgc}.

\subsection{Hybrid Method}
Hybrid approaches combine explicit context coding with implicit refinement. Xu et al.~\cite{c35_fast_inr_refinement} employ context-based residual coding with INR-based refinement. AnyPCC~\cite{c14_anypcc} proposed an Instance-Adaptive Fine-Tuning (IAFT) strategy that fine-tunes a small subset of pretrained weights for out-of-distribution instances, reducing the encoding time and model overhead of pure INR codecs. Existing hybrid solutions, however, still provide limited control over which parameters should adapt and often compress updated weights as post-processing. Hybrid-PCGC therefore shifts the hybrid design from fixed fine-tuning to rate-controlled adaptive-parameter coding, where the updated model capacity and its transmitted bitrate are optimized together.

\section{Preliminary}
\label{sec:preliminary}

\subsection{Adaptation Unit}
\label{sec:adaptation_unit}
A set of point clouds serving as an adaptation unit is defined as $\mathcal{A}=\{x_1, \dots, x_t, \dots,x_M\}$, where $M$ is the length of the set. Each $x_t$ at index $t$ is composed of $\{C_t, F_t\}$, representing the coordinates of occupied points ($C_t$) and their associated attributes ($F_t$), respectively. We define an adaptation unit as the basic unit for instance-specific refinement. Point clouds within the same adaptation unit have no interdependent relationship and are used only to share the instance adaptation time and network parameters. In the following sections, we use $x_t$ as the basic unit for the description of the network architecture, and abbreviate it as $x$ for notational convenience.

\subsection{Bottom-Up Octree Construction}
\begin{figure}[htp]
    \centering
    \includegraphics[width=0.95\linewidth]{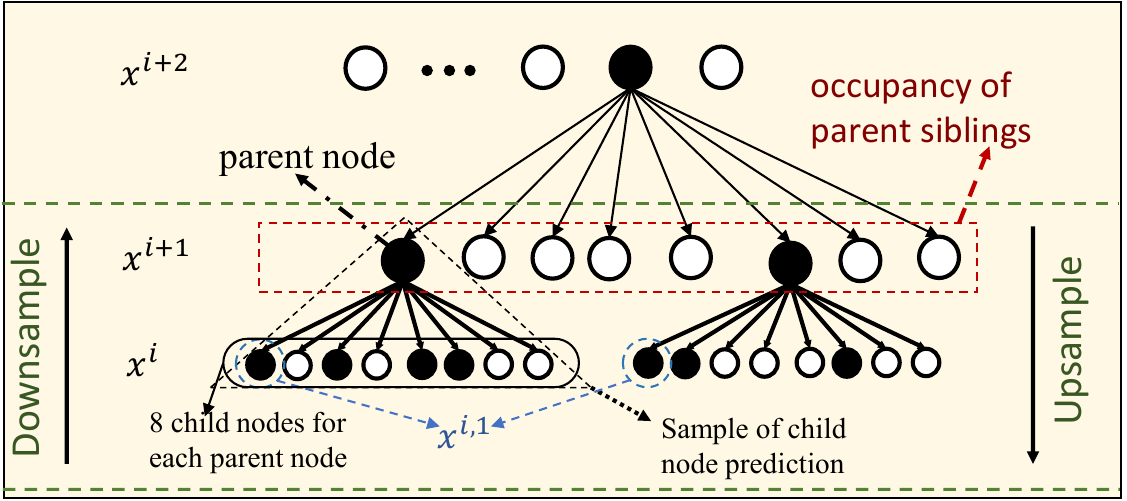}
    \caption{Concept in octree.}
    \label{fig:octree}
    \vspace{-1em}
\end{figure}
The bottom-up octree construction shown in Fig. \ref{fig:octree} is a core data structure for most point cloud codecs. Each node is associated with a single-bit occupancy code, 0 for non-occupied and 1 for occupied. We define the octree levels as spatial scales. Downsample from a high-scale $x^i$ (child nodes) to a low-scale $x^{i+1}$ (parent nodes) is functionally equivalent to a Max Pooling operation with a $2^3$ kernel. $x^{i,j}$ denotes the $j$-th child node of all nodes with octant $j$ in scale $i$ (e.g., the blue line in Fig. \ref{fig:octree} represents all the first child nodes, which are marked as $x^{i,1}$).  

Most learning-based PCGC methods first apply this Downsampling process until a coarse representation at scale $L$. Their main contribution lies in the subsequent use of Upsampling that predicts the occupancy status of each child node from its parent nodes scale-by-scale. The predicted probabilities of child nodes from parent nodes are then leveraged by an arithmetic coder to encode/decode the actual occupancy. Finally, non-occupied nodes are pruned to generate a lossless reconstruction of a higher scale.  The main task of our work is to improve the Upsampling accuracy for a better prediction of higher scale point clouds from lower scale point clouds, resulting in a better compression performance.

\begin{figure*}[t]
  \centering
  \includegraphics[width=\textwidth]{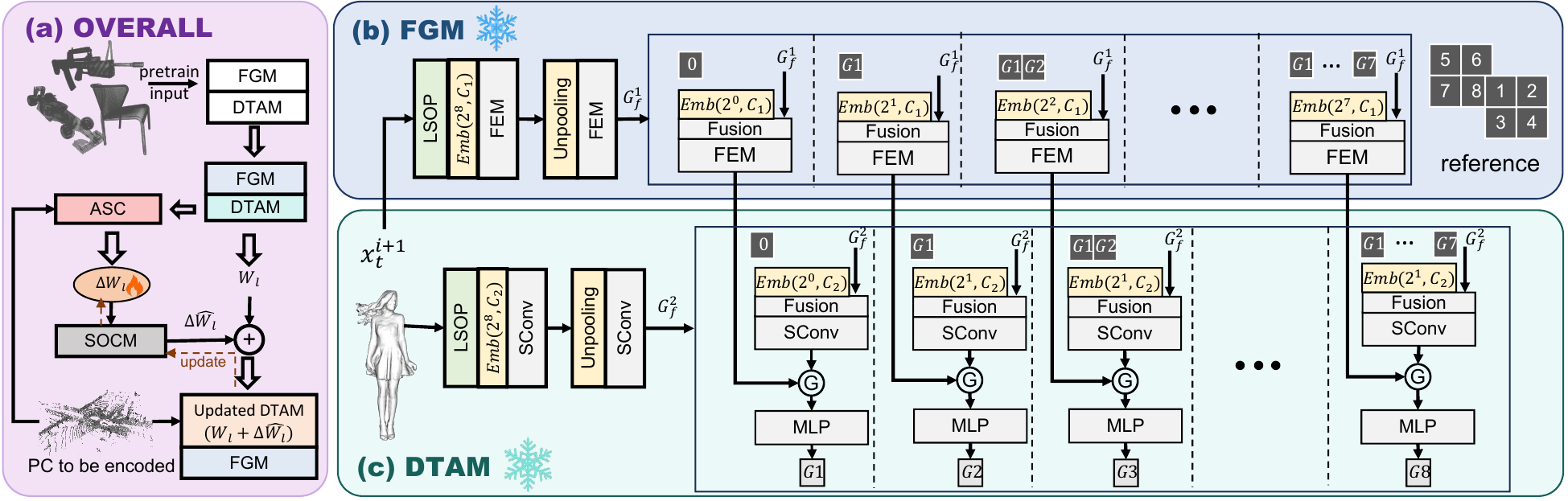}
    \caption{(a) The framework includes the Pipeline. (b) The architecture of the FGM, which iteratively generates a robust prior. (c) The architecture of the DTAM, which consumes the prior from the FGM to refine the final occupancy prediction $P_{occ}^{i+1,j}$.}
  \label{fig:framework}
  \vspace{-1.5em}
\end{figure*}

\section{Method}
\label{sec:method}

\subsection{Overall}
The Hybrid-PCGC framework integrates a pretrained model with an INR to achieve efficient distribution-agnostic point cloud geometry compression. As shown in Fig. \ref{fig:framework} (a), the pipeline can be divided into three steps:
\begin{itemize}
    \item Offline prior extraction. Use a large-scale training dataset to train the FGM and DTAM, which provides a robust basis.

    \item Instance Adaptation. Given the target point cloud, ASC first determines the subset of DTAM parameters that should participate in instance adaptation, denoted as $W_l \subset W_{\mathrm{DTAM}}$. We then instantiate a learnable offset $\Delta W_l$ with the same tensor structure as $W_l$, while keeping both FGM and the original DTAM weights frozen. During instance adaptation, the selected DTAM parameters are re-parameterized as
    \begin{align}
        \widetilde{W}_l = W_l + \Delta W_l ,
    \end{align}
    and $\widetilde{W}_l$ is used for DTAM inference. Only the offset $\Delta W_l$ is optimized on the target point cloud, allowing the model to adapt to instance-specific geometric statistics without modifying the pretrained weights. Meanwhile, the SOCM is jointly optimized to supervise the bitrate of $\Delta W_l$, balancing the occupancy bitrate gain and the additional model overhead.  

    \item Compression. Entropy-code the geometry with the predictions from FGM and the adapted DTAM. Instead of transmitting the entire DTAM, only the selected adaptation scope $l$ and the quantized offset $\Delta \widehat{W}_l$ are compressed into the bitstream. The decoder reconstructs the effective DTAM parameters as $W_l+\Delta \widehat{W}_l$ for geometry decoding.

\end{itemize}
\subsection{Frozen General Module (FGM)}
\begin{figure}[htp]
    \centering
    \includegraphics[width=0.85\linewidth]{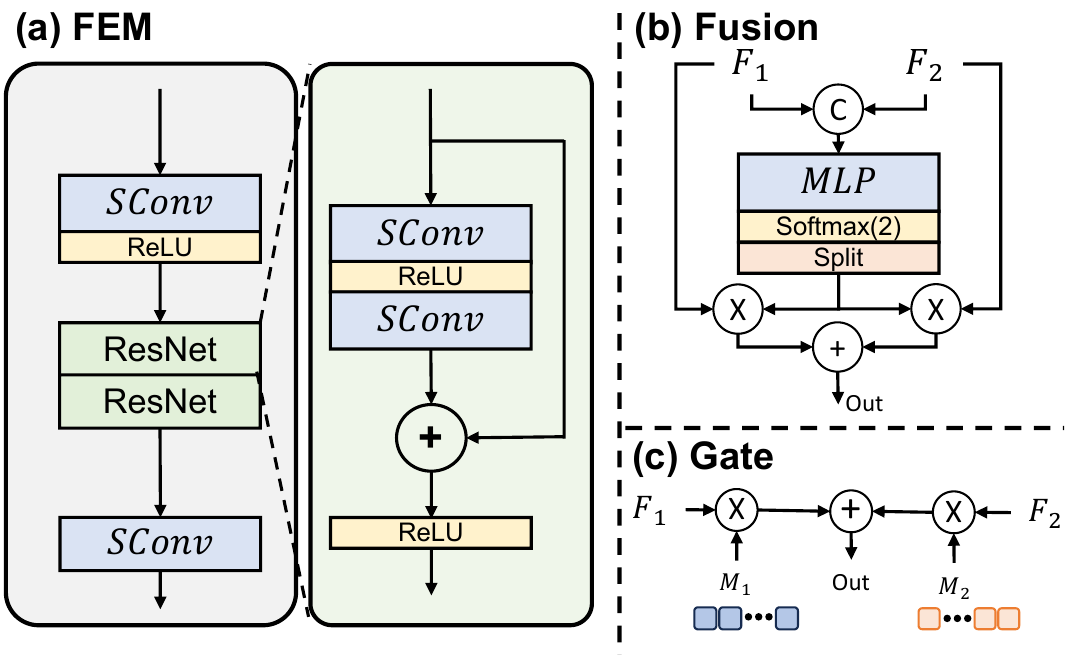}
    \caption{Component of FGM. (a) FGM and ResNet. (b) Fusion module. It is used to integrate global information with the decoded local information. (c) Gate. It employs learnable weighting coefficients $M_1$ and $M_2$ to control the weighted sum of the two input features.}
    \label{fig:FGM component}
    \vspace{-1em}
\end{figure}
As illustrated in Fig. \ref{fig:framework} (b), the FGM serves as the architectural cornerstone of our framework, engineered to extract robust and dataset-agnostic prior knowledge that accelerates the convergence of the subsequent adaptive module. By encapsulating a large parameter capacity within a rate-efficient architecture, this module provides a highly generalized representation foundation; moreover, since its weights remain entirely frozen and are excluded from the final bitstream, it can focus exclusively on optimizing feature extraction performance.

The prior feature derived from FGM is divided into $8$ stages. This interaction forms a core iterative loop for each stage, indexed by $j=1,2,...,8$: when decoding the $j$-th stage, FGM first generates a prior probability $Pr_1^{i,j}$ based on available information and provides it to DTAM. DTAM utilizes this prior feature to decode the child node occupancy $x^{i,j}$ for the current stage. DTAM then feeds the newly decoded child node occupancy result $x^{i,j}$ back to FGM, which uses it as updated information. The key effect of this 8-stage interaction is that the prior is progressively refined, as FGM continuously incorporates the newly decoded occupancy information from DTAM to generate a more accurate prediction for the next stage, until all child nodes at the current scale are decoded. 

The input of FGM contains two parts: 1) global input: the low-scale point cloud $x^{i+1}$, and 2) stage input: decoded child node occupancy $x^{i,<j}$ returned by DTAM. Firstly, FGM uses the $x^{i+1}$ to derive the global feature 
\begin{align}\label{eq:FGM}
    &F_{G1}^{i} = FEM(Unpooling(FEM(Embed(LSOP(x_t^{i+1}))))),
\end{align}
where Low Scale Occupancy Prior (LSOP) is a non-learning feature extraction module. It extracts the occupancy of the parent siblings as the conditional feature to predict the occupancy of the child nodes. The siblings of the parent node are marked in red rectangles in Fig. \ref{fig:octree}. The Feature Extraction Module (FEM) depicted in Fig. \ref{fig:FGM component} (a) is a block used to extract features. $SConv(\cdot)$ and $ResNet(\cdot)$ denote Sparse Convolution and Residual Network. Subsequently, the decoded child nodes can be used to extract context embedding $C_{FGM}^{i,j}$.
\begin{align}
    &C_{FGM}^{i,j} = \begin{cases}
        Embed_j(\textbf{0}), &j=1, \nonumber \\
        Embed_j(bin2dec_j(x^{i,<j})), & otherwise,
    \end{cases}
\end{align}
where $x^{i,<j}$ indicates that the child nodes decoded in the previous $j-1$ stages are directly concatenated together, which can be denoted as $\mathrm{Concat}(x^{i,1}, x^{i,2}, \cdots, x^{i,j-1})$. The $bin2dec_j$ denotes the conversion of j-bit binary numbers to decimal. And the $Embed_j$ is a learnable embedding specifically designed to supply context information for the $j$-th stage. Finally, FGM calculates the prior information $Pr_1^{i,j}$ from $C^{i,j}$ and $F_G^{FGM}$,
\begin{align}
    Pr_1^{i,j} = FEM(Fusion(C_{FGM}^{i,j}, F_{G1}^{i})), 
\end{align}
where $Fusion$ is a fusion module depicted in Fig. \ref{fig:FGM component} (b).

\subsection{Dual Trainable Adaptive Module (DTAM)}
DTAM depicted in Fig. \ref{fig:framework} (c) is designed to perform online instance adaptation while simultaneously ensuring rapid convergence and reducing the parameter bitrate. This is achieved through a refining mechanism that builds upon the prior information from FGM.

While DTAM also operates in 8 stages per scale, its internal feature propagation method is deliberately different from FGM. The input data of DTAM contains three parts: 1) low-scale point cloud $x^{i+1}$, 2) decoded child node occupancy $x^{i,<j}$, 3) the prior $Pr_1^{i,j}$ derived from FGM. Firstly, we use the same process as FGM to obtain the feature $Pr_2^{i,j}$ from $x^{i+1}$ and the decoded child node occupancy $x^{i,<j}$,
\begin{align}
    &F_{G2}^{i} = SConv(Unpooling(SConv(Embed(LSOP(x^{i+1}))))), \\
    &C_{DTAM}^{i,j} = \begin{cases}
        Embed_j(0), &j=1, \nonumber \\
        Embed_j(bin2dec_j(x^{i,<j})), & otherwise,
    \end{cases} \\
    &Pr_2^{i,j} = SConv(Fusion(C_{DTAM}^{i,j}, F_{G2}^{i})),
\end{align}
where $F_{G2}^i$ represents the global features shared in all stages at scale $i$. The $C_{DTAM}^{i,j}$ denotes stage context embedding in the DTAM, which is derived from decoded child node occupancy. The embedding parameters here are not shared with those in FGM. The feature $Pr_2^{i,j}$ represents the prior information independently extracted by DTAM. Subsequently, we use a gate to determine the fusion weight of $Pr_1^{i,j}$ and $Pr_2^{i,j}$
\begin{align}
    T^{i,j} = Gate(Pr_1^{i,j},Pr_2^{i,j}),
\end{align}
where $Gate$ depicted in Fig. \ref{fig:FGM component} (c) is a gate controller that performs channel-wise weighted addition. $T^{i,j}$ is the fused feature that has both general information and instance information. Next, we use an MLP network to project $T^{i,j}$ into the child node occupancy $P_{occ}^{i,j}$, 
\begin{align}
    &P_{occ}^{i,j} = SoftMax(MLP(T^{i,j})),
\end{align}
where the MLP outputs two logits and $SoftMax$ normalizes them into the probabilities of non-occupied and occupied symbols, respectively. Although the occupied probability is mathematically equivalent to applying a sigmoid function to the difference between the two logits, two-class softmax provides more stable optimization in sparse occupancy prediction. The resulting $P_{occ}^{i,j}$ is used by the arithmetic coder as the binary symbol probability for child-node occupancy. Finally, for the convenience of bit rate supervision during training, we use binary cross-entropy $Bce$ to estimate the bit stream size in each stage,
\begin{align}
    &B^{i,j} = Bce(x^{i,j},P_{occ}^{i,j}), \nonumber \\
    &Bce(u,v)=-(u\log_2(v)+(1-u)log_2(1-v)).
\end{align}

\subsection{Adaptive Scope Controller (ASC)}






ASC is proposed to determine which subset of DTAM parameters should participate in instance adaptation. Instead of using a fixed update scope for all inputs, ASC assigns an adaptive parameter scope according to the coding difficulty of the target adaptation unit.

Given a target adaptation unit $\mathcal{A}$, ASC first evaluates its coding difficulty using the bitrate estimated by the frozen FGM-DTAM pipeline before instance adaptation. Intuitively, a larger non-adapted bitrate indicates that the pretrained prior is less matched to the current geometry and that a larger adaptive capacity is required. We therefore use the non-adapted geometry bitrate as the selection score,
\begin{align}
    R_0(\mathcal{A}) = \sum_{x \in \mathcal{A}} \sum_i \sum_j B^{i,j}(x),
\end{align}
where $B^{i,j}(x)$ denotes the estimated occupancy bitrate at the $j$-th stage of scale $i$, computed by the frozen FGM-DTAM pipeline \textbf{without} instance adaptation. The score jointly reflects the number of occupied nodes and the prediction uncertainty of the frozen model. ASC then assigns the adaptation scope by comparing $R_0(\mathcal{A})$ with a small set of predefined thresholds:
\begin{align}
    l =
    \begin{cases}
        -1, & R_0(\mathcal{A}) < \tau_1, \\
        0, & \tau_1 \le R_0(\mathcal{A}) < \tau_2, \\
        1, & \tau_2 \le R_0(\mathcal{A}) < \tau_3, \\
        2, & \tau_3 \le R_0(\mathcal{A}) < \tau_4, \\
        3, & R_0(\mathcal{A}) \ge \tau_4 ,
    \end{cases}
\end{align}
where $l$ is the adaptive scope index and specifies how many DTAM components are assigned instance-specific offsets. The thresholds $\{\tau_1,\tau_2,\tau_3,\tau_4\}$ are empirically set according to the bitrate range observed on representative validation samples and are kept fixed for all test point clouds. They serve as fixed empirical hyperparameters rather than instance-optimized variables. In this way, simple point clouds can skip adaptation or update only a very small parameter subset, while complex point clouds are allocated a larger adaptive capacity. Each scope corresponds to a predefined subset of DTAM parameters:
\begin{align}
    W_l = \mathcal{S}_l(W_{\mathrm{DTAM}}), \quad l \in \{-1,0,1,2,3\}.
\end{align}
Here, $\mathcal{S}_l(\cdot)$ denotes the deterministic scope-selection mapping that extracts the corresponding DTAM parameter subset according to the following scope definition: $l=-1$ disables instance adaptation and denotes no parameter subset; $l=0$ denotes only the final linear layer of the occupancy-prediction MLP; $l=1$ extends the update scope to the MLPs involved in each DTAM stage; $l=2$ further includes the stage-wise SConv layers preceding these MLPs; and $l=3$ includes the entire DTAM parameter set. For the selected subset $W_l$, we create a learnable offset $\Delta W_l$ with the same tensor structure and construct the effective parameters as
\begin{align}
    \widetilde{W}_l = W_l + \Delta W_l .
\end{align}
During instance adaptation, $W_l$ remains frozen and only $\Delta W_l$ is optimized. This offset parameterization preserves the pretrained prior in DTAM while allowing the model to correct instance-specific geometric statistics with a compact set of transmittable parameters. Most test samples are not close to scope-switching boundaries, and neighboring scopes often have similar total rates near a boundary: a larger scope can reduce occupancy bits but also increase model-offset bits. Therefore, ASC mainly improves robustness by avoiding clearly under- or over-sized adaptation, while instance adaptation and SOCM provide the dominant bitrate reduction.

\subsection{Supervised Offset Compression Module (SOCM)}

\begin{figure}[htp]
    \centering
    \includegraphics[width=0.85\linewidth]{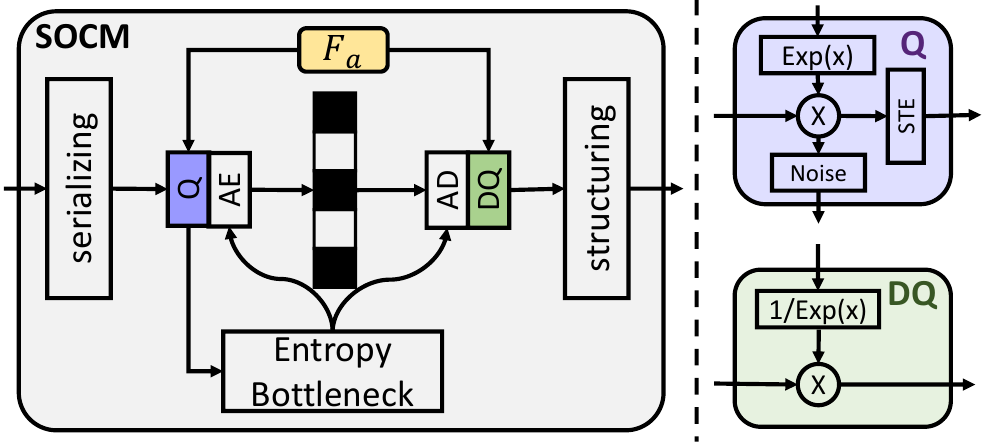}
    \caption{Principle of SOCM, where selected offset weights are quantized through STE and entropy-regularized by an entropy bottleneck. AE is an arithmetic encoder, AD is an arithmetic decoder.}
    \label{fig:SOCM}
\end{figure}

The selected offset weights are optimized online and constitute the model overhead that must be included in the bitstream. There is a trade-off: a larger adaptive scope can improve prediction accuracy while increasing the model bitrate. The SOCM module shown in Fig. \ref{fig:SOCM} is specifically introduced to address this problem. By treating the selected offset as a compressible latent variable, SOCM supervises its bitrate during instance adaptation and prevents the benefits of adaptation from resulting in prohibitive model overhead. For notational simplicity, we denote the selected offset $\Delta W_l$ as $p_m$. SOCM first vectorizes $p_m$ and applies STE-based quantization \cite{STE} to obtain the quantized offset used for adapted DTAM inference:
\begin{align}
    \hat{p}_m =
    Struct\left(
    \frac{STE(e^{F_a}\cdot Vec(p_m))}{e^{F_a}}
    \right),
\end{align}
where $F_a$ is a learnable parameter associated with quantization, and $Vec(\cdot)$ and $Struct(\cdot)$ denote vectorization and inverse restructuring, respectively. Meanwhile, an entropy bottleneck \cite{VAE} models the probability distribution of the quantized offset and estimates its bitrate:
\begin{align}
    \tilde{q}_m &= e^{F_a}\cdot Vec(p_m) + U(-0.5,0.5), \nonumber \\
    Bit_m &= -\sum \log_2 P_{\theta}(\tilde{q}_m),
\end{align}
where $\tilde{q}_m$ is the standard differentiable proxy for quantization during training, and $P_{\theta}(\cdot)$ is estimated by the entropy bottleneck. The estimated bitrate $Bit_m$ is then used to regularize the selected offset $p_m$, i.e., $\Delta W_l$.

\subsection{Lossy Extension Compatibility}
Although the proposed framework is designed for lossless geometry compression, we also report lossy results for compatibility with existing benchmark protocols. Lossy coding follows the highest-resolution lossy evaluation protocol adopted by SparsePCGC and AnyPCC~\cite{c6_sparse_tensor_pcc,c14_anypcc}: the coarse and middle scales use lossless coding with our probability model, while the finest lossy layers select the top-$k$ candidate child nodes according to predicted probabilities to match the transmitted target point count. This top-$k$ selection and target-count matching rule is inherited from prior evaluation practice~\cite{c6_sparse_tensor_pcc,c14_anypcc} and is not claimed as a methodological contribution of this paper. The lossy experiments are therefore used only to verify that the proposed lossless probability model and instance-adaptive offsets remain compatible with this established lossy evaluation setting.

\subsection{Training Strategy}
The proposed framework is trained in three steps. The first two steps are offline training stages for learning a strong base model, while the last step is performed online for each target adaptation unit.

In the first step, we train FGM to provide an accurate general prior. Specifically, the contribution of $Pr_2^{i,j}$ in Gate is set to zero, so that the occupancy prediction is driven by the FGM prior $Pr_1^{i,j}$. The objective only contains the estimated occupancy bitrate:
\begin{align}
    \mathcal{L}_{\mathrm{geo}}(x)
    =
    \frac{1}{N_x}
    \sum_i \sum_j B^{i,j}(x),
    \label{eq:offline_training_loss}
\end{align}
where $N_x$ is the number of points in the training sub-cloud $x$, and $B^{i,j}(x)$ denotes the estimated bitrate of the occupancy symbols at the $j$-th stage of scale $i$.

In the second step, the parameters of FGM are fixed, and Gate is enabled for end-to-end fusion training. This stage allows the general prior $Pr_1^{i,j}$ and the DTAM prior $Pr_2^{i,j}$ to be sufficiently integrated before online adaptation. The same loss in Eq. \ref{eq:offline_training_loss} is used, and no model bitrate term is introduced because no instance-specific offset is transmitted during offline base-model training.

In the third step, online instance adaptation is conducted for the target adaptation unit $\mathcal{A}=\{x_t\}_{t=1}^{T}$. ASC first determines the adaptive scope $W_l$, and only the corresponding offset $\Delta W_l$ is optimized while the base model is frozen. In this stage, the objective includes both the point-cloud bitrate and the model bitrate supervised by SOCM. The point cloud bitrate is accumulated over all point clouds in the adaptation unit, all spatial scales, and all stages:
\begin{align}
    \mathcal{L}_{\mathrm{pc}}(\mathcal{A})
    =
    \frac{1}{N_{\mathcal{A}}}
    \sum_{x_t\in\mathcal{A}} \sum_i \sum_j B^{i,j}(x_t),
    \label{eq:adaptation_pc_loss}
\end{align}
where $N_{\mathcal{A}}=\sum_{x_t\in\mathcal{A}}N_{x_t}$ is the total number of points in the adaptation unit. The model bitrate is normalized in the same bpp form:
\begin{align}
    \mathcal{L}_{\mathrm{model}}(\mathcal{A})
    =
    \frac{Bit_m}{N_{\mathcal{A}}}.
    \label{eq:adaptation_model_loss}
\end{align}
The final online adaptation objective is therefore
\begin{align}
    \mathcal{L}_{\mathrm{IA}}(\mathcal{A})
    =
    \mathcal{L}_{\mathrm{pc}}(\mathcal{A})
    +
    \mathcal{L}_{\mathrm{model}}(\mathcal{A}).
    \label{eq:online_adaptation_loss}
\end{align}
After optimization, the selected adaptive scope and the compressed offset weights are encoded into the bitstream together with the occupancy symbols.
\section{Experiment}
\subsection{Experiment Setup}

\textbf{Datasets.} We follow the benchmark protocol of AnyPCC~\cite{c14_anypcc} to evaluate lossless point cloud geometry compression on diverse data sources. The test set covers standard dynamic human sequences, including 8iVFB~\cite{8ivfb}, MVUB~\cite{mvub}, Owlii~\cite{owlii}, and Thuman~\cite{c60_thuman_function4d}; indoor scenes and LiDAR scans, including ScanNet~\cite{c59_scannet}, KITTI~\cite{c57_kitti}, and Ford~\cite{c58_ford}; and more challenging point clouds from large-scale reconstruction and neural rendering pipelines. Following AnyPCC, we also include Dense, Sparse, GS, VGGT, S3DIS, RS, NS, and CS to examine robustness under broader and out-of-distribution cases~\cite{c14_anypcc,c61_3d_gaussian_splatting,c62_vggt,c63_gauspcc,c64_dl3dv}. The variants marked as Ours and Ours-U correspond to dataset-specific and unified training settings, respectively. Ours follows the conventional generalization-based protocol by training separate models for different data categories, while Ours-U trains a single model from mixed data across all categories. The detailed category split and checkpoint selection rules are provided in the supplementary material.

\textbf{Baselines.} The comparison includes both traditional and learning-based geometry codecs. G-PCC is used as the MPEG anchor, implemented with the latest public TMC13 v23 reference software~\cite{gpcc_software,GPCC_method}. For learned codecs, we compare with SparsePCGC~\cite{c6_sparse_tensor_pcc}, Unicorn~\cite{c7_umc_geometry}, OctAttention~\cite{c9_octattention}, and TopNet~\cite{c13_topnet}. We further include AnyPCC and AnyPCC-U~\cite{c14_anypcc}, which are closely related hybrid universal models. To keep results comparable, the testing protocol and reported baselines are aligned with the AnyPCC benchmark whenever possible, and detailed dataset definitions and codec settings are provided in the supplementary material.

\textbf{Metrics.} The primary metric is bits per point (bpp), which measures the total number of encoded bits normalized by the number of input points. We also report the compression-ratio gain (CR Gain) relative to the G-PCC anchor, along with the encoding/decoding time and the total number of model parameters. These metrics jointly reflect compression efficiency, computational cost, and model complexity. Rate-distortion curves provide a complementary visualization of compression performance across representative datasets.

\textbf{Implementation.} Our method is implemented with PyTorch and sparse tensor operations. All runtime measurements are obtained on the same machine for fairness. The hardware platform runs Ubuntu 22.04.5 LTS on an x86\_64 system with dual AMD EPYC 7Y43 CPUs, 192 logical CPU cores, and a single NVIDIA RTX 4090 GPU. The adaptation unit length $M$ determines the number of points participating in instance adaptation, and thus affects the rate–distortion performance. Since existing methods are generally equivalent to the $M=1$ case in testing, we conduct the main experiments under this setting. The cases with $M>1$ are discussed separately in the next experiments.

\subsection{Lossless Compression}
\begin{table*}[t]
\centering
\small
\setlength{\tabcolsep}{5pt}
\begin{tabular}{l|ccccccc|cc}
\toprule
\textbf{Dataset} & \textbf{SparsePCGC} & \textbf{Unicorn} & \textbf{OctAttention} & \textbf{TopNet} & \textbf{GPCC} 
&  \textbf{AnyPCC}&\textbf{AnyPCC-U}& \textbf{Ours}& \textbf{Ours-U}\\ 
\midrule
\textbf{8iVFB}   & 0.57 & 0.57 & 0.68 & 0.59 & 0.76 
&  \second{0.54} &0.57 
& \best{0.52}
& \second{0.54}
\\
\textbf{MVUB}    & 0.69 & 0.69 & 0.76 & 0.69 & 0.94 
&  \second{0.67} &0.75 
& \best{0.65}
& 0.70
\\
\textbf{Owlii}   & 0.48 & 0.48 & 0.62 & 0.56 & 0.59 
&  0.47 &0.47 
& \second{0.45}& \best{0.44}
\\
\textbf{Thuman}  & 1.70 & 1.70 & 2.31 & 2.20 & 2.00 
&  \second{1.58} &1.64 
& \best{1.56}
& 1.59
\\
\textbf{ScanNet} & 1.86 & 1.86 & 2.13 & 2.03 & 2.03 
&  1.83 &1.88 
& \best{1.78}& \second{1.81}
\\
\textbf{KITTI}   & 6.80 & 6.50 & 7.21 & 6.85 & 8.19 
&  \second{6.18} &6.45 
& \best{6.03}
& 6.35
\\
\textbf{Ford}    & 9.77 & 8.44 & 9.10 & 8.54 & 10.32 
&  \second{8.40} &8.57 
& \best{8.36}
& 8.54
\\ 
\cmidrule(lr){1-10}
\textbf{Dense}   & 6.37 & 5.48 & 6.55 & 6.38 & 5.32 
&  \second{5.27} &5.55 
& \best{4.62}
& 5.47
\\
\textbf{Sparse}  & 9.98 & 9.42 & 10.40 & 10.02 & 9.35 
&  \second{9.11} &9.26 
& \best{8.89}& 9.20\\
\textbf{GS}      & 15.82 & / & \second{11.31} & \best{10.95} & 14.46 
&  11.65 &11.74 
& 11.61
& 11.7
\\
\textbf{VGGT}    & 7.84 & / & 8.22 & 7.83 & 7.33 
&  7.30 &\second{7.06} 
& 7.32
& \best{6.83}
\\
\textbf{S3DIS}   & 11.88 & / & 11.52 & 10.84 & \second{10.66} 
&  10.93 &10.79 
& 10.77
& \best{10.59}
\\ 
\cmidrule(lr){1-10}
\textbf{RS}      & 3.88 & / & 4.05 & 3.92 & 3.72 
&  3.68 &\second{3.50} 
& 3.65
& \best{3.41}
\\
\textbf{NS}      & 6.54 & / & 4.89 & 4.84 & 4.85 
&  4.69 &\second{4.67} 
& 4.69
& \best{4.64}
\\
\textbf{CS}      & 4.94 & / & 3.40 & 3.21 & 3.23 
&  3.18 &\second{3.08} 
& 3.18& \best{3.00}\\ 
\midrule
\textbf{CR Gain $\downarrow$} & 2.07\% & / & 1.32\% & -4.04\% & 0.00\% 
&  \second{-11.93\%} &-10.75\% 
& \best{-13.96\%} & -13.07\%\\
\textbf{Enc/Dec (s) $\downarrow$} & 2.6/2.2 & / & 7.7/1324 & 8.7/1740 & 3.8/2.7 &  2.84/0.46 & 2.84/0.46 &  4.51/1.68 & 4.51/1.68   \\
\bottomrule
\end{tabular}
\caption{Lossless geometry compression performance on the benchmark. Bpp is reported for each dataset, with the best value in each row highlighted with a \textcolor{red}{red} background and the second-best value highlighted with a \textcolor{blue}{blue} background. CR Gain is computed relative to G-PCC, and Enc/Dec denotes average encoding/decoding time. Ours and Ours-U correspond to the dataset-specific and unified settings, respectively. The results for Unicorn are cited directly from the original publication.}

\label{tab:main_result}
\vspace{-1em}
\end{table*}

\noindent\textbf{Bitstream performance.} Table~\ref{tab:main_result} reports the lossless geometry compression results. Overall, the proposed method achieves the best average compression performance among the compared codecs. Compared with the MPEG G-PCC anchor, ours obtains a -13.96\% CR Gain. It also outperforms representative learning-based codecs, including SparsePCGC and Unicorn from the sparse-tensor and multiscale coding family, as well as OctAttention and TopNet from the octree-based context modeling family. These results indicate that the proposed hybrid design is effective across diverse categories of point clouds rather than being tailored to a specific data source.

Among the compared methods, AnyPCC is the most recent and strongest hybrid baseline, reporting state-of-the-art performance. Ours achieves a larger average CR gain from -11.93\% to -13.96\%. Notably, the proposed method achieves more visible bitrate reductions on several representative challenging datasets. For example, on Dense, the bitrate is reduced from 5.27 bpp to 4.62 bpp, and on KITTI, it is reduced from 6.18 bpp to 6.03 bpp. The unified setting further highlights the robustness of the proposed framework. Ours-U achieves -13.07\% CR Gain, outperforming AnyPCC-U (-10.75\%) and even surpassing the category-specific AnyPCC model (-11.93\%). In particular, Ours-U improves over AnyPCC-U on VGGT, S3DIS, RS, NS, and CS, showing better stability under out-of-distribution and reconstruction-derived inputs.

\noindent\textbf{Runtime performance.} Compared with AnyPCC, our encoding/decoding time increases from 2.84/0.46 s to 4.51/1.68 s due to adaptive scope selection and offset optimization. This runtime remains practical and is far lower than autoregressive octree codecs such as OctAttention and TopNet.

\subsection{Lossy Compression}
\begin{figure*}[t]
    \centering
     \includegraphics[width=0.95\linewidth]{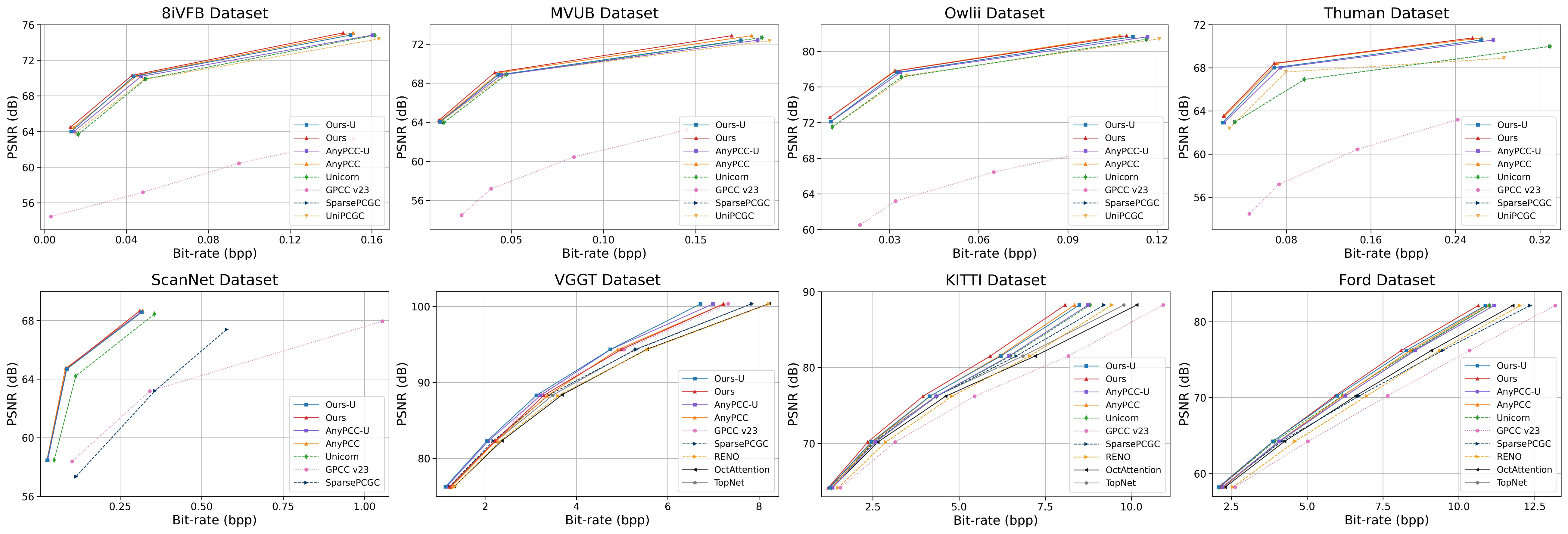}
    \caption{Performance comparison using rate-distortion curves. Comparisons on the human-body and ScanNet datasets exclude RENO, OctAttention, and TopNet, as they are specifically designed for lossy LiDAR compression and thus incompatible.}
    \label{fig:rdcurves}
\end{figure*}

\begin{table}[t]
\centering
\scriptsize
\setlength{\tabcolsep}{5pt}
\begin{tabular}{l|ccccc}
\toprule
\textbf{Dataset} & \textbf{UniPCGC} & \textbf{AnyPCC} & \textbf{AnyPCC-U} & \textbf{Ours} & \textbf{Ours-U} \\
\midrule
8iVFB   & 2.96\%& -15.01\%& -6.74\%& -19.00\%& -13.43\%\\
MVUB    & 3.64\%& -10.49\%& 0.04\%& -16.14\%& -4.29\%\\
Owlii   & 1.50\%& -20.45\%& -10.72\%& -19.26\%& -13.96\%\\
Thuman  & -16.88\%& -50.41\%& -40.43\%& -52.23\%& -44.04\%\\
ScanNet & /     & -75.88\%& -74.88\%& -75.54\%& -74.89\%\\
\midrule
Avg.    & -2.20\%& -34.45\%& -26.54\%& -36.44\%& -30.12\%\\
\bottomrule
\end{tabular}
\caption{BD-Rate comparison on dense object and indoor point clouds using SparsePCGC as the anchor. Negative values indicate bitrate savings over SparsePCGC, and lower values are better.}
\label{tab:bd_rate_dense}
\end{table}

\begin{table}[t]
\centering
\scriptsize
\setlength{\tabcolsep}{4pt}
\begin{tabular}{l|rrrrrr}
\toprule
\textbf{Dataset} & \textbf{RENO} & \textbf{TopNet} & \textbf{AnyPCC} & \textbf{AnyPCC-U} & \textbf{Ours} & \textbf{Ours-U} \\
\midrule
KITTI & -12.31\%& -16.81\%& -23.37\%& -20.07\%& -26.83\%& -23.19\%\\
Ford  & -8.84\%& -17.55\%& -18.60\%& -17.47\%& -21.72\%& -20.64\%\\
VGGT  & 9.75\%& 4.77\%& -0.42\%& -4.88\%& -0.43\%& -6.31\%\\
\midrule
Avg.  & -3.80\%& -9.86\%& -14.13\%& -14.14\%& -16.33
\%& -16.72\%\\
\bottomrule
\end{tabular}
\caption{BD-Rate comparison on LiDAR and reconstruction-derived point clouds using G-PCCv23 as the anchor. Negative values indicate bitrate savings over G-PCCv23, and lower values are better.}
\label{tab:bd_rate_lidar}
\end{table}

Fig.~\ref{fig:rdcurves} shows rate-distortion (RD) curves of lossy compression, where curves closer to the upper-left region indicate higher reconstruction quality at the same bitrate or fewer bits at the same reconstruction quality. And Table~\ref{tab:bd_rate_dense} and Table~\ref{tab:bd_rate_lidar} further provide quantitative BD-Rate results. As G-PCC performs substantially worse than the learned codecs and has limited overlap with other curves, SparsePCGC is adopted as the anchor for BD-Rate calculation. 

Under the same top-$k$ target-count matching rule as the prior lossy evaluation setting~\cite{c6_sparse_tensor_pcc,c14_anypcc}, Ours and Ours-U generally obtain competitive RD curves on dense object-level point clouds and indoor scenes. Compared with AnyPCC, the difference mainly comes from the underlying probability model and instance-adaptive offset optimization used before the inherited lossy selection step. Thus, the lossy comparison supports the compatibility of our lossless-oriented hybrid model with the standard lossy benchmark setting.

For sparse LiDAR and reconstruction-derived point clouds, existing methods do not implement lossy compression using the top-$k$ target-count matching rule. Instead, they adjust the bitrate by pre-quantizing the point coordinates. Under this setting, our gains are also clear. On KITTI and Ford, Ours achieves more favorable RD curves than LiDAR-oriented or octree-based methods such as RENO, OctAttention, and TopNet, while also improving over AnyPCC. On VGGT, the unified model Ours-U shows a clear advantage over AnyPCC-U, suggesting that the proposed framework better handles reconstruction-based point clouds that deviate from standard training distributions. It is also worth noting that Ours and Ours-U generally provide RD gains over their corresponding AnyPCC counterparts. Since Ours-U uses a single unified model without category-specific model selection, its improvement over AnyPCC-U on sparse and reconstruction-derived datasets demonstrates the generalization ability of the codec under heterogeneous lossy compression scenarios.

\subsection{Adaptation Unit Size Analysis}
As discussed in Sec. \ref{sec:adaptation_unit}, the default testing protocol follows existing codecs and sets $M=1$, where each point cloud is independently encoded and decoded. However, in practical scenarios, multiple point clouds with similar geometry distributions are often processed together. In this case, we can set $M>1$ and group several point clouds into one adaptation unit, so that they share the same adapted weight offset $\Delta W_l$. This design does not introduce dependency among the decoded point clouds, but amortizes the bitrate overhead of the transmitted network offset over more points.

\begin{table}
    \centering
    \begin{tabular}{c|cc|c}
    \toprule
             &  Ours-U& Ours-U.M & CR-Gain\\
    \midrule
         Ford &  8.57 & 8.51 & -0.70\% \\
         Dense & 5.55 & 5.45 & -1.80\% \\
         RS &  3.50 & 3.4 & -2.86\% \\
         NS &  4.69 & 4.62 & -1.49\% \\
         CS&  3.18 & 2.96 & -6.92\% \\
    \midrule
         Avg. &    5.098&      4.988&      -2.16\% \\
    \bottomrule
    \end{tabular}
    \caption{Effect of using $M>1$ point clouds in one adaptation unit. Ours-U denotes the unified model with $M=1$, while Ours-U.M shares one adapted offset among multiple point clouds.}
    \label{tab:MultiFrame}
\end{table}

Table~\ref{tab:MultiFrame} compares the default unified setting Ours-U with Ours-U.M, where $M$ is allowed to be larger than 1. In this experiment, the adaptation scope is directly set to the largest level, i.e., $l=3$, so that the grouped point clouds share the full DTAM offset. Compared with the $M=1$ setting, Ours-U.M consistently reduces the bitrate across all tested datasets. The average bpp decreases from 5.10 to 4.99, corresponding to an additional 2.16\% bitrate reduction. The gain is especially clear on CS, where sharing the offset reduces the bpp from 3.18 to 2.96.

These results demonstrate that enlarging the adaptation unit is an effective way to reduce the relative cost of instance-adaptive network parameters when the grouped point clouds have similar distributions. Therefore, although the main experiments use $M=1$ for a fair comparison with existing methods, the proposed framework can further improve practical coding efficiency by jointly adapting multiple related point clouds.

\subsection{Comparison of Basic Model Performance}

To further separate the contribution of the generalized base model from online instance adaptation, we evaluate the compression performance without instance adaptation. In this setting, no weight offset $\Delta W_l$ is optimized or transmitted, and the bitrate is only determined by the pretrained unified model. We select four representative datasets with different geometric characteristics and coding difficulties, including 8iVFB, Ford, GS, and CS.

\begin{table}
    \centering
    \begin{tabular}{c|ccc}
    \toprule
         Dataset &   G-PCC & AnyPCC-U.B & Ours-U.B \\
    \midrule
         8iVFB &   0.57& 0.56& 0.54  \\
         Ford &   9.77& 8.57& 8.55  \\
         GS &   15.82& 13.29& 13.27  \\
         CS &   4.94& 3.36& 3.28\\
    \midrule
         Avg. &   7.78& 6.45& 6.41
\\
         CR-Gain &   0.00\%& -17.11\%& -17.56\%\\
    \bottomrule
    \end{tabular}
    \caption{Comparison of unified base models without instance adaptation. AnyPCC-U.B and Ours-U.B denote the corresponding base models before optimizing instance-specific weight offsets.}
    \label{tab:Basic_Compress}
\end{table}

Table~\ref{tab:Basic_Compress} compares our unified base model, denoted as Ours-U.B, with the unified base model of AnyPCC, denoted as AnyPCC-U.B. Ours-U.B achieves a lower bpp on all tested datasets. The consistent gains across human-body, LiDAR, Gaussian-splatting-derived, and challenging reconstruction-derived point clouds indicate that the proposed base architecture provides a more robust generalized prior. On average, Ours-U.B reduces the bitrate from 6.45 bpp to 6.41 bpp compared with AnyPCC-U.B, and improves the CR Gain from -17.11\% to -17.56\% relative to G-PCC.

These results show that the proposed base model is not intended to be the sole source of the final gain. Its role is to provide a compact and robust generalized prior, while preserving a structurally separated DTAM space for controlled instance adaptation. The base-model improvement over AnyPCC-U.B is modest but consistent; the larger gain of Hybrid-PCGC comes from the complete rate-controlled system, where the base prior, adaptive scope selection, and entropy-aware offset coding jointly balance geometry bits and model-offset bits.

\subsection{Weight Offset Distribution Analysis}

\begin{figure}[htpb]
    \centering
     \includegraphics[width=0.85\linewidth]{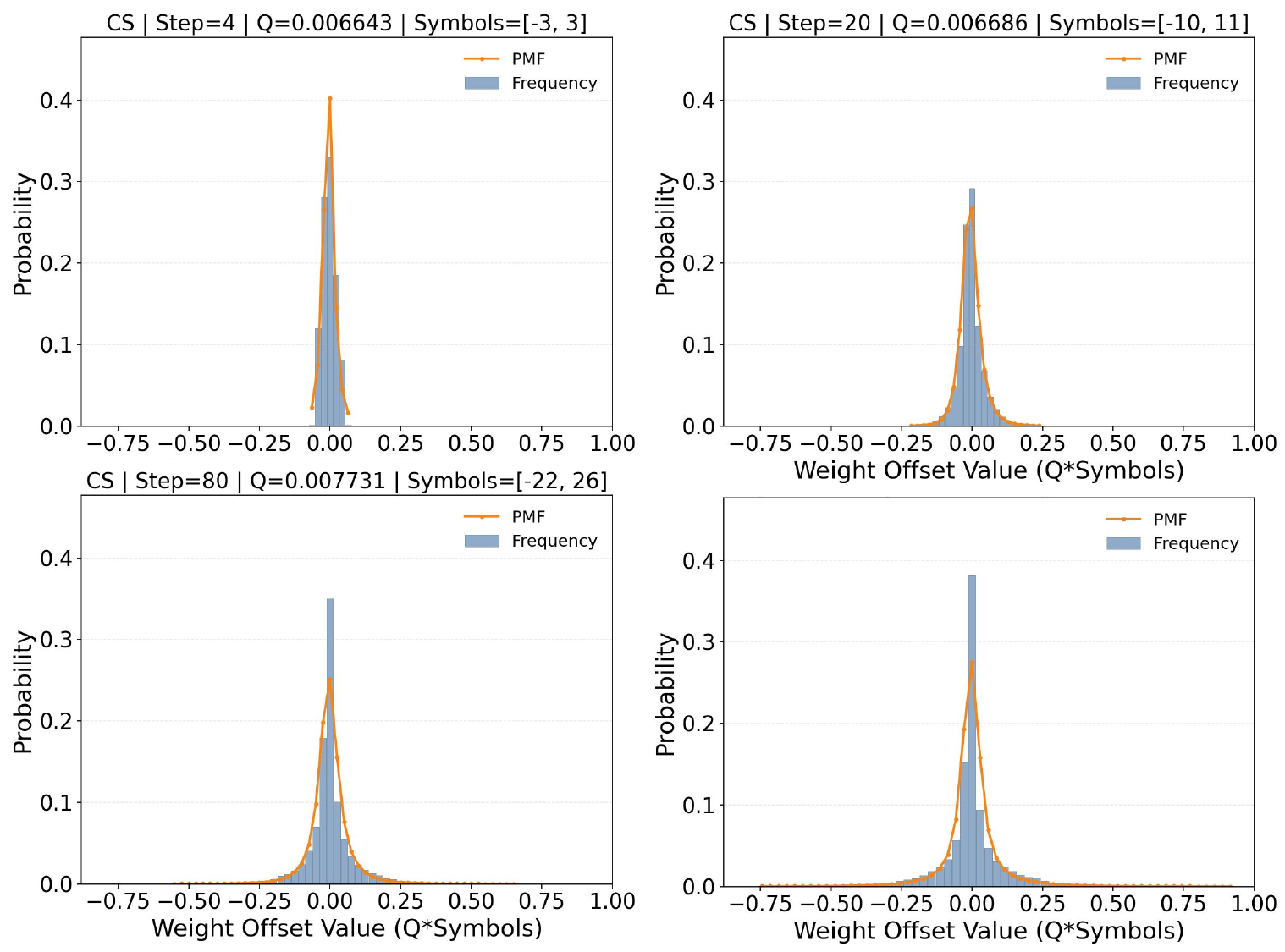}
    \caption{Distribution evolution of transmitted weight offsets on CS during online instance adaptation. The blue histogram denotes the empirical frequency of quantized offsets, and the orange curve denotes the PMF estimated by SOCM.}
    \label{fig:distribution}
    \vspace{-1em}
\end{figure}

Fig.~\ref{fig:distribution} visualizes the distribution evolution of the transmitted weight offsets during online instance adaptation. We show four optimization steps, including 4, 20, 80, and 160. The blue histogram represents the empirical frequency of the quantized weight offsets, while the orange curve denotes the probability mass function estimated by the SOCM entropy model. The bin width corresponds to the learned quantization step size $Q$, and the horizontal axis represents the reconstructed offset value, i.e., $Q$ multiplied by the quantized symbols.

As adaptation proceeds, the quantized symbol range gradually changes, and the learned quantization step is adjusted accordingly. This indicates that SOCM does not use a fixed parameter coding scale, but adaptively controls the quantization of transmitted offsets according to the evolving weight distribution. Meanwhile, the offset distribution remains highly concentrated around zero, especially in later optimization steps. This suggests that instance adaptation mainly performs compact parameter correction: most weights require only small changes, while a limited number of offsets capture instance-specific geometric statistics.

More importantly, the estimated PMF consistently matches the empirical offset histogram across different optimization steps. Even when the offset distribution becomes sharper near zero or develops wider tails, SOCM can still provide an accurate probability estimation. This verifies that the proposed SOCM module effectively models the transmitted weight offsets and provides reliable entropy supervision for controlling the model bitrate.

\subsection{Bpp Allocation}

\begin{figure}[htpb]
    \centering
     \includegraphics[width=0.85\linewidth]{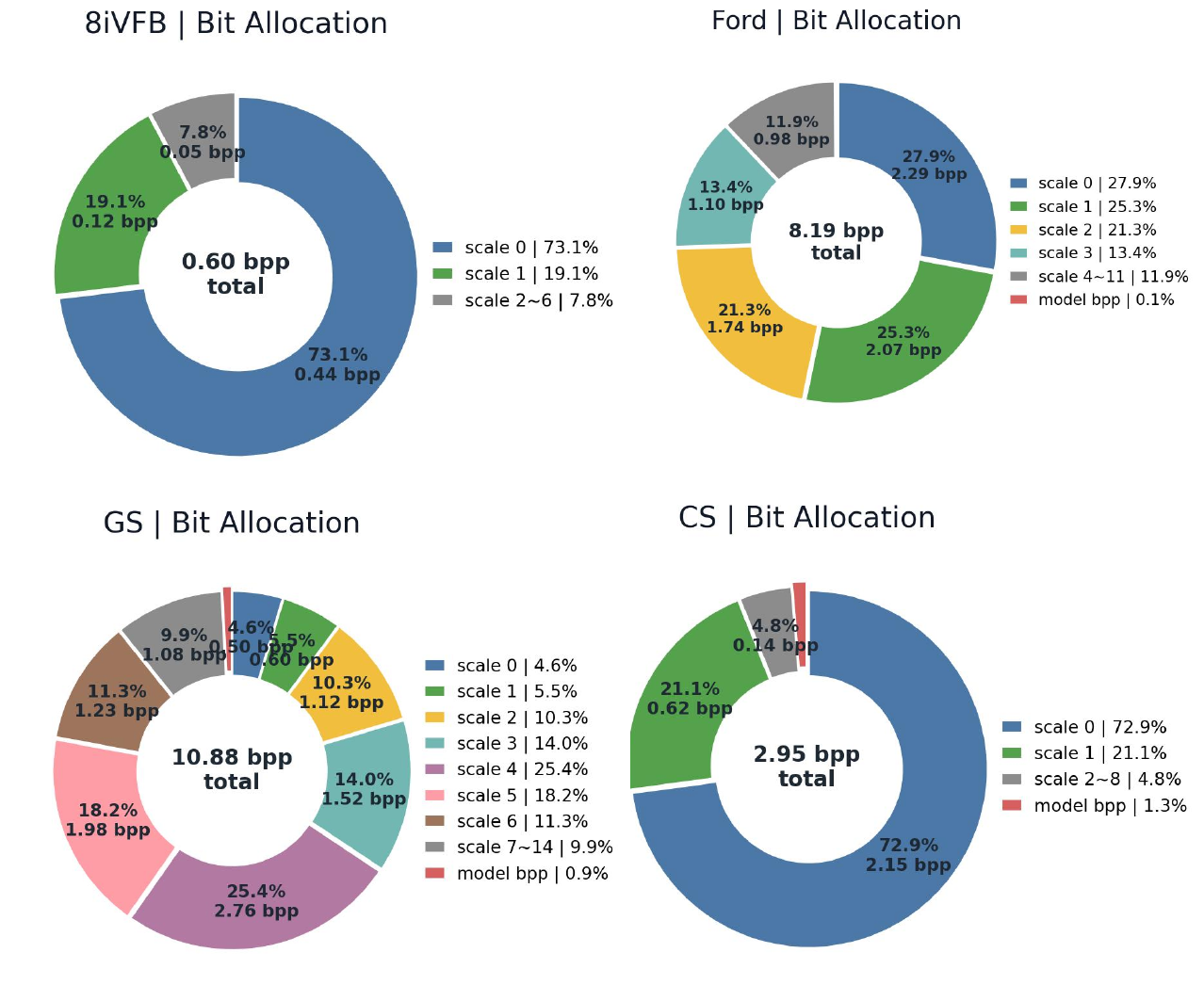}
    \caption{Bitrate allocation across geometry scales and transmitted model offsets on representative datasets.}
    \label{fig:allocation}
    \vspace{-1em}
\end{figure}

Fig.~\ref{fig:allocation} further analyzes the bitrate allocation of the proposed codec across geometry scales and transmitted model offsets. We visualize four representative datasets with different geometric characteristics, including 8iVFB, Ford, GS, and CS. Each donut chart reports the total bpp and the relative bitrate contributions from different geometry scales and model bpp. Here, model bpp corresponds to the compressed weight offsets used to calibrate the pretrained network for the target point cloud. They also include the side information required for decoding the model-offset stream, such as the quantized entropy-bottleneck parameters and the quantized $\exp(F_a)$, which are transmitted in the actual bitstream.

The first observation is that the proportion of model bpp varies across datasets. For data where the pretrained prior requires stronger instance-specific correction, the codec allocates a larger bitrate to the transmitted weight offsets. This behavior reflects the adaptive nature of our framework: the model does not assign a fixed overhead to all inputs, but adjusts the parameter bitrate according to the coding difficulty and the distribution mismatch of the target point cloud. This also explains why datasets with a relatively higher model-bit proportion can benefit more from the $M>1$ setting discussed in the adaptation unit analysis, where the same offset is shared by multiple similar point clouds and its overhead is amortized.

The second observation is that the geometry bitrate is distributed differently across spatial scales for different types of point clouds. For dense point clouds such as 8iVFB, a large portion of the bitrate is concentrated at high-resolution scales, where fine geometric details need to be represented. For sparse LiDAR point clouds such as Ford, the bitrate is spread more evenly over multiple scales, and the contribution of each scale is relatively moderate. By contrast, GS exhibits a distinct allocation pattern, where the dominant bitrate appears in intermediate scales rather than following a purely dense or sparse pattern. CS also shows its own allocation preference, reflecting the distribution difference of reconstruction-derived point clouds.

These results show that Hybrid-PCGC jointly balances scale-wise geometry coding and model-offset coding across diverse distributions.

\subsection{Ablation}

We conduct ablation studies to isolate the contribution of each instance-adaptive component in the proposed unified model. Ours-U denotes the complete setting, where online instance adaptation is enabled, ASC selects the adaptive scope, and SOCM optimizes the transmitted weight offsets under entropy supervision. Abla.1 removes ASC and uses a fixed adaptation scope for all inputs. We report the best fixed setting with $l=1$. Abla.2 further removes SOCM. In this case, the model parameters are directly fine-tuned instead of optimizing entropy-supervised offsets, and the resulting parameter changes are compressed by CABAC~\cite{c57_CABAC} as a post-processing step. Abla.3 disables overfitting, i.e., no instance adaptation is performed, and only the base model is used. The gain reported in this section is a within-ablation relative bitrate reduction on the four representative datasets, computed with Abla.3 as the reference, rather than the benchmark-level CR-Gain in Table~\ref{tab:main_result}. The component configurations are summarized in Table~\ref{tab:ablation}.

\begin{table}
    \centering
    \begin{tabular}{c|ccc|c}
        \toprule
                 &  IA  &  SOCM   &   ASC  & Gain vs. Abla.3 \\
        \midrule
        Ours-U     & \cmark & \cmark & \cmark & -7.21\%\\
        Abla.1& \cmark & \cmark & \xmark &  -5.93\%\\
        Abla.2& \cmark & \xmark & \xmark &  -5.66\%\\
        Abla.3& \xmark & \xmark & \xmark &  0.00\% \\
        \bottomrule
    \end{tabular}
    \caption{Component ablation of the proposed unified model. IA denotes online instance adaptation. Gain is computed within this ablation study using Abla.3 as the reference.}
    \label{tab:ablation}
\end{table}

\begin{table}
    \centering
    \begin{tabular}{c|cccl}
    \toprule
         Dataset&  Ours-U&  Abla.1&  Abla.2&Abla.3\\
    \midrule
         8iVFB&  0.54&  0.56&  0.54&0.54
\\
         Ford&  8.54&  8.66&  8.55&8.55
\\
         GS&  11.7&  11.7&  12.14&13.27
\\
         CS&  3.00&  3.00&  3.10&3.61
\\
    \midrule
         Gain vs. Abla.3&  -7.21\%&  -5.93\%&  -5.66\%&0.00\%\\
    \bottomrule
    \end{tabular}
    \caption{Detailed bitrate results for the ablation study. Gain is computed over the four representative datasets using Abla.3 as the reference.}
    \label{tab:ablation_detail}
\end{table}

The detailed results in Table~\ref{tab:ablation_detail} show a clear performance degradation as each component is removed. Comparing Abla.3 with Abla.2, enabling instance adaptation improves the within-ablation gain from 0.00\% to -5.66\%. This demonstrates that instance adaptation is the main source of improvement, since it can correct the distribution mismatch between the unified base model and the target point cloud. The gain is especially visible on GS and CS, where the base model alone is less matched to the test distribution.

Comparing Abla.2 with Abla.1 further verifies the importance of SOCM. Without SOCM, the parameter update is optimized independently of its final coding cost and then compressed by CABAC as a separate post-processing step. Introducing SOCM improves the within-ablation gain from -5.66\% to -5.93\%, showing that entropy-aware offset optimization provides a better rate trade-off than directly fine-tuning network parameters and compressing them afterward. Finally, Ours-U improves over Abla.1 by replacing the fixed scope $l=1$ with ASC. The gain increases from -5.93\% to -7.21\%, indicating that adaptive scope selection helps avoid the capacity mismatch caused by using the same update range for point clouds with different coding difficulties.

Overall, the complete Ours-U configuration achieves the best within-ablation gain of -7.21\%. These results confirm that the gain is not obtained by merely fine-tuning a pretrained codec. Instead, instance adaptation, SOCM, and ASC form a closed rate-control loop: instance adaptation provides the correction capability, SOCM controls the transmitted parameter cost, and ASC further improves robustness by selecting a suitable adaptation scope.

\section{Conclusion}
In this paper, we propose Hybrid-PCGC, a hybrid point cloud geometry compression framework that treats instance-specific model offsets as rate-constrained transmitted variables. By decoupling FGM-DTAM, ASC, and SOCM, the proposed method jointly controls geometry bitrate and model-offset bitrate. Experiments on diverse point cloud datasets demonstrate that Hybrid-PCGC provides robust compression performance across heterogeneous data sources.

\bibliographystyle{IEEEtran}
\bibliography{main} 

\section{Details of Parameters}
The main training and online instance-adaptation parameters used in our experiments are listed in \cref{tab:parameters}. Following the training strategy in the main paper, Stage I trains FGM to learn a general prior, Stage II freezes FGM and trains the DTAM-Gate fusion branch, and the online stage optimizes only the selected adaptive offsets for each target adaptation unit.

\begin{table}[h]
\centering
\small
\begin{tabularx}{\linewidth}{c|X|c}
\toprule
\textbf{Symbol} & \textbf{Description} & \textbf{Value} \\
\midrule
$C_{\mathrm{FGM}}$ & Network channels used for FGM prior modeling in Stage I & 64 \\
$K_{\mathrm{FGM}}$ & Convolution kernel size used for FGM prior modeling in Stage I & 5 \\
$C_{\mathrm{DTAM}}$ & Network channels used for DTAM refinement in Stage II & 8 \\
$K_{\mathrm{DTAM}}$ & Convolution kernel size used for DTAM refinement in Stage II & 5 \\
$B$ & Batch size & 1 \\
$lr_{\mathrm{off}}$ & Initial learning rate for offline training & 0.0005 \\
$\gamma_{\mathrm{off}}$ & Learning-rate decay factor for offline training & 0.1 \\
$S_{\mathrm{I}}$ & Maximum training steps in Stage I & 17e4 \\
$D_{\mathrm{I}}$ & Learning-rate decay steps in Stage I & 10e4, 15e4 \\
$S_{\mathrm{II}}$ & Maximum training steps in Stage II & 15e4 \\
$D_{\mathrm{II}}$ & Learning-rate decay steps in Stage II & 8e4, 12e4 \\
$S_{\mathrm{IA}}$ &  Maximum training steps in instance adaptation& 50 or 200 \\ 
$lr_{\mathrm{IA}}$ & Initial learning rate for online instance adaptation & 0.004 \\
$\gamma_{\mathrm{IA}}$ & Learning-rate decay factor for online instance adaptation & 0.995 \\
$T_{\mathrm{IA}}$ & Learning-rate decay interval for online instance adaptation & 1 \\
$Q_{\mathrm{init}}$ & Initial quantization step size for transmitted weight offsets & $1/150$ \\
$G_{\mathrm{script}}$ & Default GOP size in the testing script & 10 \\
$F_{\mathrm{num}}$ & Number of frames or point clouds to be encoded; $-1$ denotes all inputs & -1 \\
$s_{\mathrm{lossy}}$ & Lossy geometry scale, set to 0 for the lossless setting & 0 \\
$u_{\mathrm{AE}}$ & Whether to use arithmetic encoding in the instance-adaptation script & 0 \\
\bottomrule
\end{tabularx}
\caption{Training and online instance-adaptation parameters used in our experiments. Stage I, Stage II, and online instance adaptation correspond to the three-step training strategy described in the main paper. Dataset-dependent online settings follow the lossless testing protocol described in the main paper.}
\label{tab:parameters}
\end{table}

In all offline training experiments, we use data augmentation and patch-based training. Stage II uses the pretrained FGM from Stage I as the frozen general foundation and trains the lightweight adaptive branch, so that the subsequent online instance adaptation can start from a stable generalized prior. In the online stage, the dataset-dependent parameters are limited to the ASC-selected adaptation level and the number of instance-adaptation steps used by the lossless testing protocol. In single-instance testing, level $3$ is generally not triggered and is mainly reserved for the $M>1$ adaptation-unit setting.

\section{Benchmark and Codec Configuration}
\label{sec:supp_benchmark}
Our lossless evaluation follows the AnyPCC benchmark protocol. Dense, Sparse, GS, VGGT, S3DIS, RS, NS, and CS use the same data arrangement, preprocessing, and test split as the AnyPCC benchmark. The test-set statistics are summarized in \cref{tab:supp_dataset_protocol}. The training and test sequences are kept disjoint following the AnyPCC protocol. The OOD labels follow the setting used in the main paper: standard human-body, indoor, LiDAR, Dense, Sparse, and GS datasets are treated as in-distribution for the corresponding category-specific models, while VGGT, S3DIS, RS, NS, and CS are used to evaluate out-of-distribution robustness.

\begin{table*}[t]
\centering
\begin{tabularx}{\textwidth}{lccclcl}
\toprule
\textbf{Dataset} & \textbf{Depth} & \textbf{Frames/Seq.} & \textbf{Total} & \textbf{Cond.} & \textbf{Avg. Points} & \textbf{Sequence/Source} \\
\midrule
8iVFB & 10-bit & 300 & 600 & Easy & 760375 & Loot, Redandblack \\
MVUB & 10-bit & / & 461 & Easy & 1259792 & Phil, Ricardo \\
Owlii & 11-bit & 600 & 2400 & Easy & 2635971 & All sequences \\
Thuman & 10-bit & / & 20 & Easy & 441319 & Frames 0500--0519 \\
ScanNet & 2 cm & 100 & 100 & Easy & 126760 & ScanNet test split \\
KITTI & 1 mm & 100 & 1100 & Easy & 119826 & Sequences 11--21 \\
Ford & 1 mm & 100 & 200 & Easy & 83061 & Sequences 02--03 \\
Dense & 12-bit & / & 2 & Medium & 3222415 & MPEG AI-PCC Dense vox12 \\
Sparse & 12-bit & / & 3 & Medium & 996846 & MPEG AI-PCC Sparse vox12 \\
GS & 1 mm & / & 13 & Medium & 288413 & 3D Gaussian splatting-derived point clouds \\
VGGT & 1 mm & / & 50 & Medium & 257467 & VGGT reconstruction-derived point clouds \\
S3DIS & 16-bit & / & 1 & Medium & 47062002 & Stanford Area 2 \\
RS & 10--12 bit & / & 16 & Hard & 1058013 & Random-sampling perturbation \\
NS & 10--12 bit & / & 15 & Hard & 1670503 & Noise-added perturbation \\
CS & 10--12 bit & / & 14 & Hard & 1531537 & Complex-sinusoidal deformation \\
\bottomrule
\end{tabularx}
\caption{Test dataset protocol inherited from the AnyPCC benchmark. ``Total'' denotes the number of tested point clouds/frames. ``/'' means that the dataset is not organized by a fixed frames-per-sequence value in the benchmark table. Average point counts are reported before entropy coding under the benchmark preprocessing.}
\label{tab:supp_dataset_protocol}
\end{table*}

For KITTI and Ford, the original coordinates are first quantized to 1 mm precision and then scaled by the benchmark quantization parameter $Q=16$ before lossless geometry coding. For all other datasets, the benchmark-specified coordinates or bit depths in \cref{tab:supp_dataset_protocol} are used directly. The Dense and Sparse sets correspond to the MPEG AI-PCC vox12 test sets. VGGT and GS are reconstruction-derived point clouds, while RS, NS, and CS are generated from source point clouds using controlled random sampling, additive noise, and non-rigid sinusoidal deformation, respectively.

For the traditional anchor, we use the public TMC13 v23 implementation of G-PCC and follow the MPEG common test conditions for lossless geometry coding. The anchor is evaluated in octree-based geometry coding mode; trisoup and inter prediction are not used in this static/single-frame lossless benchmark. The coordinate precision and position quantization follow each dataset definition in \cref{tab:supp_dataset_protocol}. For learned baselines, the results of SparsePCGC, Unicorn, OctAttention, TopNet, AnyPCC, and AnyPCC-U follow the AnyPCC benchmark when they are inherited in the main comparison, while Ours and Ours-U are evaluated on the same test sets. Missing entries in the main result table indicate results that were not reported under the same dataset/protocol in AnyPCC; we leave them unavailable instead of mixing benchmark-inherited results with non-aligned reimplementations.

\subsection{Dataset-Specific and Unified Settings}
The variants denoted by Ours and Ours-U differ only in the offline checkpoint used before online instance adaptation. Ours follows the category-specific protocol commonly used by previous learned PCC methods: a separate FGM/DTAM checkpoint is trained for each data category, and the checkpoint is selected according to the dataset type specified by the benchmark configuration. This setting is reported to keep the comparison fair with prior methods that also rely on category-specific training and testing protocols. Ours-U uses a single unified checkpoint for all tested categories. Therefore, Ours-U is the more practical deployment setting because it does not require maintaining multiple category-specific models or selecting a model according to the input source, while still achieving strong performance on both in-distribution and out-of-distribution point clouds.

In both settings, the pretrained checkpoint is treated as the codec shared by the encoder and decoder, as is standard for learned compression evaluation. Thus, persistent checkpoint storage is not included in the per-point-cloud bpp. For Ours, the benchmark category determines the checkpoint before coding, so no model identifier is transmitted in the tested bitstream. In a fully self-descriptive deployment, a small model identifier would be needed for the dataset-specific setting. Ours-U avoids this issue because the decoder always uses the same universal checkpoint. The performance gap between Ours and Ours-U in the main result table therefore directly reflects the cost of using only one unified model instead of category-specific checkpoints.

\begin{table*}[t]
\centering
\begin{tabularx}{\textwidth}{l|X|X}
\toprule
\textbf{Checkpoint} & \textbf{Training data/source} & \textbf{Test use in Ours} \\
\midrule
8iVFB & 8iVFB Longdress/Soldier and MVUB Andrew/David/Sarah training sequences & 8iVFB Loot/Redandblack, MVUB Phil/Ricardo, and Owlii sequences \\
Thuman & Thuman training split, frames 0000--0499 & Thuman test frames 0500--0519 \\
ScanNet & ScanNet benchmark training scans with the same preprocessing & Disjoint ScanNet benchmark test split \\
KITTI & SemanticKITTI sequences 00--10, quantized from 1 mm coordinates & KITTI sequences 11--21 and the benchmark OOD sets mapped to the LiDAR-style checkpoint \\
Ford & Ford 01 training sequence with 1 mm quantization & Ford 02--03 test sequences \\
GS & Gaussian-splatting-derived training point clouds from the benchmark training source & Disjoint GS test set \\
Dense/Sparse & MPEG AI-PCC same-type training samples such as \textit{Frog}, \textit{Head}, \textit{ULB\_Unicorn}, and \textit{Egyptian\_mask}, following the Unicorn/MPEG protocol & Dense and Sparse MPEG AI-PCC vox12 test sets, disjoint from the training samples \\
\bottomrule
\end{tabularx}
\caption{Category-specific training and checkpoint selection used by Ours. Ours-U replaces all category-specific checkpoints in this table with one unified checkpoint.}
\label{tab:supp_ours_oursu_protocol}
\end{table*}

For Dense and Sparse, we follow the protocol adopted in Unicorn for the MPEG AI-PCC static 12-bit objects: the test set contains two Dense Object samples and three Sparse Object samples, while additional same-type samples such as \textit{Frog}, \textit{Head}, \textit{ULB\_Unicorn}, and \textit{Egyptian\_mask} are used for training. This keeps the Dense/Sparse training and testing arrangement aligned with the MPEG-oriented protocol instead of tuning on the test samples.

\section{ASC Threshold Selection}
\label{sec:supp_asc_threshold}
ASC uses the non-adapted coding cost $R_0(\mathcal{A})$ to select the adaptation scope. The thresholds are fixed to
\begin{align}
    \{\tau_1,\tau_2,\tau_3,\tau_4\}
    =
    \{5\times10^5,2\times10^6,1\times10^7,5\times10^7\}.
\end{align}
They are selected once from representative validation samples according to the range of the frozen FGM-DTAM coding cost and are not tuned on the test point clouds. The score is not normalized by the number of points because ASC controls the absolute amount of adaptive parameters and model-offset bits for an adaptation unit. A point cloud with more occupied symbols naturally requires more total occupancy bits and may need a larger adaptation capacity, even if its bpp is similar to a smaller point cloud. Thus, using total non-adapted bits lets the selection reflect both geometric uncertainty and the number of symbols to be coded.

The thresholds are intentionally coarse rather than finely optimized. Samples far from a boundary keep the same scope under moderate threshold perturbations. For samples close to a boundary, the adjacent scopes usually give similar total rates because increasing $l$ reduces the occupancy bitrate but also increases the transmitted model-offset bitrate. Therefore, ASC is used mainly to avoid clearly under-sized or over-sized adaptation scopes, rather than to perform per-sample rate optimization. This also explains why ASC provides a moderate average gain in the ablation study while improving the robustness of the complete codec across different point-cloud distributions.

\subsection{Adaptive Scope Definition}
\label{sec:supp_scope_definition}
The selected scope determines which subset of DTAM receives an instance-specific offset. Following the DTAM design in the main paper, the scopes are nested: larger scopes contain all parameters in smaller scopes and further include earlier DTAM components. The scope definitions used by ASC are listed in \cref{tab:supp_scope_definition}. Entries marked by ``--'' are implementation statistics to be filled after the corresponding measurement.

\begin{table}[t]
    \centering
    \small
    \caption{Definition of ASC adaptation scopes.}
    \label{tab:supp_scope_definition}
    \begin{tabularx}{\linewidth}{c|X|c}
        \toprule
        \textbf{Scope} & \textbf{Updated DTAM components} & \textbf{Offset params}  \\
        \midrule
        $l=-1$ & No instance adaptation; no DTAM offset is transmitted & 0  \\
        $l=0$ & Final linear layer of the occupancy-prediction MLP &  528 \\
        $l=1$ & MLPs involved in each DTAM stage & 25,616  \\
        $l=2$ & Stage-wise MLPs and their preceding SConv layers & 31,008  \\
        $l=3$ & Entire DTAM parameter set & 44,952  \\
        \bottomrule
    \end{tabularx}
\end{table}

\section{Actual Bitstream Format}
\label{sec:supp_bitstream}
This section provides the byte-level syntax used to form the final compressed stream. The differentiable bitrate terms in the main paper are used to optimize the model during online instance adaptation, but the reported bpp is computed from the actual packed stream. Therefore, the final bitrate includes the arithmetic-coded geometry occupancy stream, the arithmetic-coded model-offset stream, the adaptive scope flag, the quantized SOCM parameter side information, and all stream-length headers.
\subsection{Occupancy Probability Coding}
For each child node, the final MLP outputs two logits. Softmax converts them into a binary probability vector $P_{occ}^{i,j}=(p_0,p_1)$, where $p_0$ and $p_1$ denote the probabilities of non-occupied and occupied symbols. This is equivalent to using a sigmoid for the occupied probability $p_1$, but the two-logit softmax form is used for more stable training. During arithmetic coding, the encoder and decoder use the same causal decoding order, so all context features used to compute $P_{occ}^{i,j}$ are available on both sides.

The floating-point probabilities are converted to the integer CDF required by the arithmetic coder through a fixed deterministic routine. Given a CDF precision of $P$ bits, each probability is first clipped to avoid zero-probability bins and then converted to integer frequencies whose sum is $2^P$; the cumulative sum of these frequencies forms the CDF table. The clipping value, CDF precision, and rounding rule are fixed codec constants shared by the encoder and decoder. As a result, the occupancy stream is coded with the same quantized probabilities on both sides, and the reported geometry bitrate is measured from the generated arithmetic-coded byte stream rather than from the training loss alone.

\subsection{Bitstream Syntax and Side Information}
The final reported bitrate is computed from a decodable bitstream rather than only from the differentiable bitrate estimates used during training. For each adaptation unit, the transmitted stream contains both the geometry occupancy stream and the model-offset stream, together with all side information required by the decoder. The adaptive scope $l$ selected by ASC is encoded as a 3-bit flag and is included in the model-offset bitrate. Since the selected scope uniquely determines the list of DTAM tensors to be updated, the selected offset $\Delta W_l$ is serialized by flattening these tensors into a one-dimensional sequence in a fixed layer and parameter order. Therefore, no additional variable tensor shape needs to be signaled.

For SOCM decoding, the entropy bottleneck parameters used to build the CDF are not assumed to be available implicitly at the decoder. They are quantized to 8-bit integer values and transmitted before the arithmetic-coded offset symbols. The quantizer range, scale, zero point, tensor shape, and serialization order of these parameters are fixed by the SOCM architecture and model version, so they are shared by the encoder and decoder rather than transmitted per point cloud. After dequantizing these parameters, the decoder computes the same discrete CDF as the encoder using the fixed arithmetic-coder precision and tail-handling rule. The CDF itself is not transmitted in the bitstream. The scale parameter $\exp(F_a)$ is transmitted separately as a 16-bit floating-point value, so that the decoder can reconstruct the same quantization scale and inverse-transform the decoded offset sequence. The offset symbols are the integer values obtained after scaling and rounding the flattened offset vector. The geometry occupancy streams, model-offset stream, entropy-bottleneck parameter side information, $\exp(F_a)$, scope flag, stream-length headers, and any byte alignment or flush bits are all counted in the final bpp.

\subsection{Model-Offset Side Information}
For each adaptation unit, ASC selects one adaptive scope $l$. We encode $l$ as a 3-bit flag, which supports the finite set of adaptation levels used in the experiments. This flag is included in the model-offset stream. Once $l$ is decoded, the decoder knows the exact subset of DTAM parameters that may receive offsets. The selected offsets $\Delta W_l$ are then serialized by traversing the selected DTAM tensors in a fixed module/layer order and flattening each tensor in row-major order. The flattened vectors are concatenated into a one-dimensional sequence. Since the tensor list is determined by $l$ and the network architecture is fixed, no extra tensor-shape field is transmitted.

The entropy bottleneck in SOCM is optimized for the target adaptation unit. To make the model-offset stream independently decodable, its parameters are transmitted as 8-bit integer values before the offset symbols. These values are dequantized by a fixed affine rule associated with each entropy-bottleneck parameter tensor. The affine ranges and zero points are part of the codec configuration and are therefore identical at the encoder and decoder. The decoder first reconstructs the entropy-bottleneck parameters and then computes the same fixed-precision CDF used by the arithmetic decoder. The CDF table is a deterministic result of this computation and is not transmitted. The learned scale parameter is transmitted separately as a 16-bit floating-point value $\exp(F_a)$ and included in the side information. After the integer offset symbols are decoded, the decoder applies the inverse scaling with the decoded $\exp(F_a)$ and restores $\Delta \widehat{W}_l$ in the fixed tensor order.

\subsection{Byte-Level Syntax}
The model-offset sub-bitstream follows the syntax in \cref{tab:supp_model_syntax}. Multi-byte scalar fields use little-endian order. The symbols in the length column denote fixed codec constants: $N_{\mathrm{EB}}$ is the number of entropy-bottleneck parameters, $N_{\Delta W_l}$ is the number of selected offset symbols implied by the decoded scope $l$, and $B_{\mathrm{AE}}$ is the byte length of the arithmetic-coded offset payload. The value of $N_{\mathrm{EB}}$, the entropy-bottleneck tensor shapes, the per-tensor 8-bit dequantization ranges, the CDF precision, and the finite support/tail handling for offset symbols are fixed by the SOCM implementation and model version.

\begin{table}[h]
\centering
\begin{tabularx}{\linewidth}{lclX}
\toprule
\textbf{Field} & \textbf{Type} & \textbf{Length} & \textbf{Decoding dependency} \\
\midrule
Scope flag $l$ & uint8 & 1 byte & Selects the DTAM offset tensor list and $N_{\Delta W_l}$ \\
EB parameter count & uint32 & 4 bytes & Sanity check against fixed $N_{\mathrm{EB}}$ \\
EB parameters & uint8 array & $N_{\mathrm{EB}}$ bytes & Dequantized by fixed per-tensor affine rules to compute the CDF \\
$\exp(F_a)$ & float16 & 2 bytes & Inverse scale for decoded offset symbols \\
Offset payload size & uint32 & 4 bytes & Number of bytes in the arithmetic-coded offset stream \\
Offset payload & byte array & $B_{\mathrm{AE}}$ bytes & Arithmetic decoded with the CDF computed from EB parameters \\
\bottomrule
\end{tabularx}
\caption{Byte-level syntax of the model-offset sub-bitstream. The CDF is computed at the decoder from the transmitted entropy-bottleneck parameters and fixed codec constants; it is not transmitted as side information.}
\label{tab:supp_model_syntax}
\end{table}

\subsection{Stream Packing Syntax}
The final stream is assembled from a list of byte strings. In our implementation, the sub-bitstreams are generated separately and packed in a fixed order: the ASC scope flag, the SOCM side-information stream, the arithmetic-coded model-offset stream, and the geometry occupancy streams. The same order is used by the decoder after it recovers the byte boundaries. The packing routine is:

{\scriptsize
\begin{verbatim}
def pack_bitstream(streams, dtype='uint32'):
    num = np.array(len(streams), dtype=dtype).tobytes()
    sizes = [len(s) for s in streams]
    header = np.array(sizes, dtype=dtype).tobytes()
    payload = b''
    for s in streams:
        assert len(s) < 2 ** 32 - 1
        payload += s
    return num + header + payload
\end{verbatim}
}

The first field stores the number of sub-bitstreams, and the next fields store the byte length of each sub-bitstream. We use \texttt{uint32} for these fields, so the header cost is $(K+1)\times32$ bits for $K$ sub-bitstreams. This header is included in the reported bitrate. After parsing the header, the decoder slices the packed stream into the original sub-bitstreams and decodes them in order.

\subsection{Decoding Order}
The decoder follows the following order:
\begin{enumerate}
    \item Parse the packed-stream header and recover all sub-bitstream boundaries.
    \item Decode the ASC scope flag $l$.
    \item Decode the 8-bit entropy-bottleneck parameter values, dequantize them by the fixed affine rules, and compute the CDF for the offset symbols.
    \item Decode the 16-bit floating-point $\exp(F_a)$.
    \item Arithmetic-decode the flattened offset sequence $\Delta \widehat{W}_l$ using the reconstructed CDF.
    \item Reshape the decoded sequence into the selected DTAM tensors according to the fixed architecture order and reconstruct the adapted weights.
    \item Decode the geometry occupancy streams using FGM and the adapted DTAM.
\end{enumerate}
Any byte alignment or flush bits introduced by the arithmetic coder are part of the corresponding byte string and are therefore counted automatically after packing. No uncounted side information is used by the decoder.


\end{document}